\documentclass[letterpaper,journal]{IEEEtran}
\usepackage{amsmath,amsfonts}
\usepackage{algorithmic}
\usepackage{algorithm}
\usepackage{array}
\usepackage[caption=false,font=normalsize,labelfont=sf,textfont=sf]{subfig}
\usepackage{textcomp}
\usepackage{stfloats}
\usepackage{url}
\usepackage{verbatim}
\usepackage{graphicx}
\usepackage[utf8]{inputenc}
\usepackage{lineno}
\usepackage[numbers,sort&compress]{natbib}
\usepackage{amsmath,amssymb,amsfonts}
\usepackage{algorithmic}
\usepackage{graphicx}
\usepackage{textcomp}
\usepackage{hyperref}
\usepackage{booktabs}
\usepackage{multirow}
\usepackage{colortbl}
\usepackage[table,xcdraw]{xcolor}
\usepackage{fancyhdr}
\usepackage{comment}
\usepackage{xspace}
\usepackage{setspace}

\begin{document}

%\title{When Statistics, Semantics, and Texture Align: A Multi-Feature Fusion Approach for GenAI Images Detection}
\title{Multi-Feature Fusion Approach for Generative AI Images Detection}
\author{Abderrezzaq Sendjasni,
        Mohamed-Chaker Larabi, ~\IEEEmembership{Senior Member,~IEEE}
        % <-this % stops a space
\thanks{This work is funded by the Nouvelle-Aquitaine Research Council under project REALISME AAPR2022-2021-17027310. \\
This work has been submitted to IEEE Transactions for possible publication. Copyright may be transferred without notice, after which this version may no longer be accessible.
}% <-this % stops a space
}

\markboth{Journal of \LaTeX\ Class Files,~Vol.~14, No.~8, August~2021}%
{Sendjasni \MakeLowercase{\textit{et al.}}: A Multi-Feature Fusion Approach for GenAI Images Detection}

% % column for its text to clear the IEEEpubid mark.

\maketitle

\begin{abstract}
The rapid evolution of Generative AI (GenAI) models has led to synthetic images of unprecedented realism, challenging traditional methods for distinguishing them from natural photographs. While existing detectors often rely on single-feature spaces, such as statistical regularities, semantic embeddings, or texture patterns, these approaches tend to lack robustness when confronted with diverse and evolving generative models. In this work, we investigate and systematically evaluate a multi-feature fusion framework that combines complementary cues from three distinct spaces: (1) Mean Subtracted Contrast Normalized (MSCN) features capturing low-level statistical deviations; (2) CLIP embeddings encoding high-level semantic coherence; and (3) Multi-scale Local Binary Patterns (MLBP) characterizing mid-level texture anomalies. Through extensive experiments on four benchmark datasets covering a wide range of generative models, we show that individual feature spaces exhibit significant performance variability across different generators. Crucially, the fusion of all three representations yields superior and more consistent performance, particularly in a challenging mixed-model scenario. Compared to state-of-the-art methods, the proposed framework yields consistently improved performance across all evaluated datasets. Overall, this work highlights the importance of hybrid representations for robust GenAI image detection and provides a principled framework for integrating complementary visual cues. 
\end{abstract}

\vspace{-0.5cm}
\section{Introduction}
\label{sec:intro}

\IEEEPARstart{R}{ecent} advances in generative artificial intelligence (GenAI) have fundamentally transformed digital content creation. Models such as StyleGAN \cite{8977347}, DALL·E \citep{pmlr-v139-ramesh21a}, Stable Diffusion \cite{Rombach_2022_CVPR}, and Midjourney are now capable to produce synthetic images that are virtually indistinguishable from natural photographs to the untrained eye. While these technologies unlock unprecedented creative potential, they also raise significant concerns regarding misinformation, copyright infringement, and the erosion of trust in digital media \cite{StealthDiffusion}. As a result, reliably distinguish natural images from AI-generated content has become a critical challenge at the intersection of multimedia forensics, computer vision, and cybersecurity.

Early approaches to this problem relied on the observation that initial generative models introduced detectable statistical anomalies. Methods based on Natural Scene Statistics (NSS), such as BRISQUE \cite{6272356} and NIQE \cite{6353522}, successfully exploited these deviations in luminance and contrast distributions. However, as generative models have advanced, they have become increasingly adept at reproducing these low-level statistical properties, thereby reducing the effectiveness of purely statistical detectors~\cite{Rombach_2022_CVPR}. This ongoing arms race between generation and detection has motivated the exploration of complementary detection paradigms.

More recent approaches have explored alternative detection strategies, including (1) \textit{semantic analysis} using vision-language models like CLIP \cite{pmlr-v139-radford21a} to identify logical inconsistencies or prompt-image mismatches \cite{Cozzolino_2024_CVPR}; (2) \textit{frequency-domain analysis} to uncover unnatural spectral patterns that may not be observable in the spatial domain inspection \cite{10334046, Li_2024_CVPRW}; and (3) \textit{texture-based methods} such as Local Binary Patterns (LBP) to capture repetitive micro-patterns and boundary artifacts characteristic of synthetic content \cite{pietikainen2010local, Durall_2020_CVPR}. While each approach has demonstrated promising results on specific models or datasets, a critical research gap remains: the lack of a comprehensive understanding of their relative strengths, complementarity, and generalization capabilities across the diverse and rapidly evolving landscape of modern GenAI models \cite{lin2024detecting, schinas2024sidbench}.

A key question remains whether any single feature space is universally superior, or whether its effectiveness depends on the underlying generative architecture? For instance, can texture-based analysis reliably detect diffusion-based model outputs that excel at reproducing natural textures? Most importantly, does a principled fusion of these orthogonal feature spaces lead to a detector that is both more robust and more model-agnostic than single-feature approaches? Addressing these questions is essential for advancing the field beyond specialized detectors toward general-purpose solutions.

In this work, we present a systematic investigation and a unified fusion framework to address these challenges. Our contributions are threefold:

-\textbf{Comprehensive benchmarking:} We conduct a rigorous evaluation of three representative and complementary feature spaces, MSCN (NSS), CLIP (semantics), and MLBP (texture), across four benchmark datasets and multiple state-of-the-art GenAI models, revealing their respective strengths and limitations.
    
-\textbf{Multi-Feature Fusion Framework:} We propose a fusion pipeline that strategically integrates statistical, semantic, and texture-based representations, demonstrating that their combination consistently outperforms single-feature baselines in terms of accuracy, robustness, and generalization.
    
-\textbf{In-depth analytical insights:} Through dimensionality reduction analysis and evaluation on challenging mixed-model scenarios, we provide empirical evidence of feature complementarity and offer practical guidelines for designing more effective GenAI detection systems.

The remainder of this article is organized as follows: Section \ref{sec:related} reviews related work on GenAI detection. Section \ref{sec:method} details the proposed methodology and feature extraction processes. Section \ref{sec:exp} describes the experimental setup, presents the results, and discusses their implications and limitations. Finally, Section \ref{sec:conclusion} concludes the article and outlines directions for future research.

\vspace{-0.1cm}
\section{Related Work}
\label{sec:related}

Research on GenAI image detection has evolved through several overlapping phases, closely following advances in generative modeling. Early methods focused on model-specific forensic analysis, while more recent approaches increasingly emphasize generalization across diverse generators through learned or handcrafted feature representations.

\textbf{Statistical and low-level feature methods:} These approaches are based on premise that generative processes introduce statistical anomalies not present in natural images. Methods such as BRISQUE \cite{6272356} and NIQE \cite{6353522} model natural scene statistics in the spatial domain, particularly through the distribution of Mean-Subtracted Contrast-Normalized (MSCN) coefficients. Extensions of this paradigm also investigate artifacts in the frequency domain, where synthetic images may exhibit unnatural spectral signatures, such as mid-frequency smoothing or high-frequency spikes \cite{ricker2022towards, 10334046}. While effective for early GAN-based models, these methods often struggle to generalize as modern diffusion and transformer-based generators increasingly replicate natural image statistics~\cite{Rombach_2022_CVPR}.

\textbf{Semantic and deep feature approaches:} The emergence of large vision-language models, particularly CLIP \cite{pmlr-v139-radford21a}, has enabled a paradigm shift toward semantic-level detection. The key insight is that synthetic images, despite their visual realism, may contain subtle semantic inconsistencies, implausible object relationships, or missmatch with the intended prompt. CLIP embeddings' success has been fully leveraged through fine-tuning~\cite{Cozzolino_2024_CVPR} or lightweight adapters \cite{Liu_2024_CVPR} to distinguish real from synthetic images by capturing such discrepancies. In addition, contrastive learning strategies have been employed to better structure the embedding space, improving separation between real and fake distributions \cite{Baraldi_2024_CoDE, zhu2024genimage}.

\textbf{Texture and mid-level representation methods:} Texture analysis remains a powerful tool, as generative models often struggle to reproduce the complex, non-repetitive, and multi-scale textures of natural scenes. Descriptors such as Multi-scale Local Binary Patterns (MLBP) capture local structural patterns and have proven effective in identifying overly regular textures or boundary artifacts characteristic of synthetic images~\cite{pietikainen2010local, 8638330}. Recent work further explored mesoscopic properties \cite{Yu_2024_CVPR} and the combination of local and global features \cite{10246417} to improve robustness against post-processing and model variability.

\textbf{Combination and hybrid methods:} To address the limitations of single-feature approaches, recent studies advocate combining multiple feature domains. For instance, \cite{10246417} integrates global semantic features from CNNs with local forensic cues, while other works fuse frequency and spatial representations \cite{Li_2024_CVPRW} or adapt multi-stream architectures operating at different image scales \cite{Yu_2024_CVPR}. Despite these advances, a systematic comparison and integration of the three fundamental feature families, \textit{i.e} statistical (NSS), semantic (VLMs), and texture (LBP), across a broad spectrum of modern generative models remains largely unexplored. Furthermore, the relative contribution and complementarity of these feature spaces are still not well understood.

Our work builds on these developments by conducting a holistic evaluation of three complementary feature families operating at different levels of visual abstraction. We not only benchmark them individually across a diverse set of modern generative models but also demonstrate that a strategic fusion leads to a more robust generalizable detector, directly addressing the key challenge of model generalization highlighted in recent studies \cite{schinas2024sidbench}.

\vspace{-0.1cm}
\section{Methodology}
\label{sec:method}

Our proposed framework is designed to assess and combine the discriminative capabilities of three orthogonal feature spaces for the binary classification task of distinguishing natural from GenAI images. The central hypothesis is that, while individual feature spaces may be effective against specific generative models, their fusion provides complementary information, leading to improved robustness and generalization across diverse generators.

\vspace{-0.1cm}
\subsection{Overview}
\begin{figure}[htbp]
    \centering
    \includegraphics[width=\linewidth]{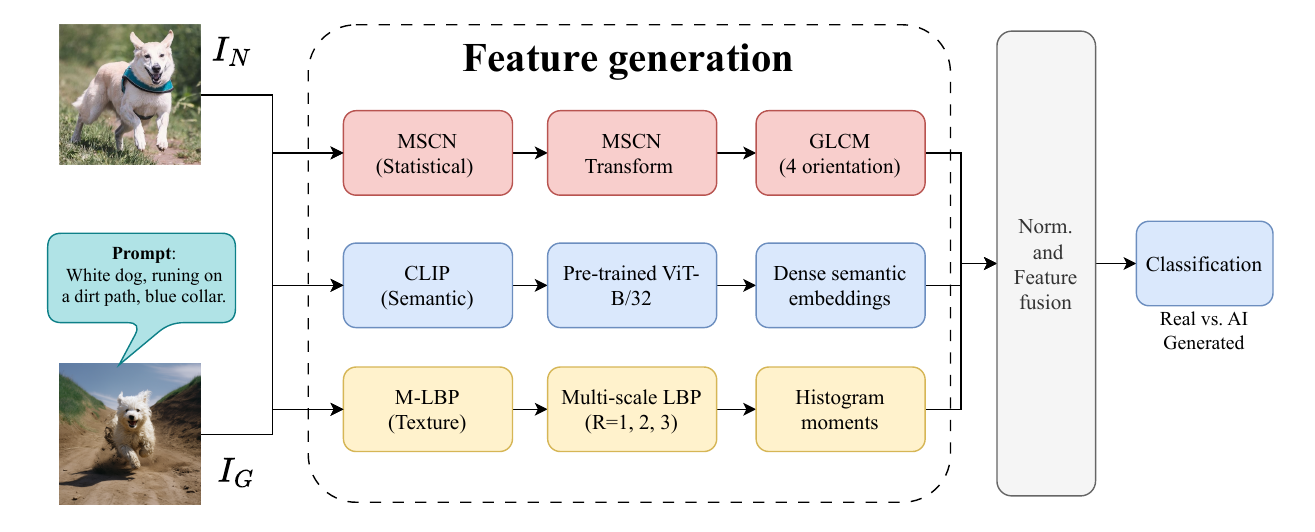}
    \caption{Overview of the proposed multi-feature fusion pipeline. An input image is processed by three parallel feature encoders (\(\Psi_{\mathrm{MSCN}}\), \(\Psi_{\mathrm{CLIP}}\), \(\Psi_{\mathrm{MLBP}}\)). The resulting feature vectors are normalized, fused, and fed into a classifier for final prediction.}
    \label{fig:framework}
\end{figure}

The overall detection pipeline, illustrated in Figure \ref{fig:framework}, consists of three main stages: (1) Parallel feature encoding, where each input image is processed by three independent feature extractors; (2) Feature normalization and fusion, where the extracted representations are standardized and fused into a unified feature vector; and (3) Classification, where a machine learning model produces the final authenticity prediction.

Formally, let $I \in \mathbb{R}^{H \times W \times 3}$ denote an input RGB image. We define three feature encoding functions:
\begin{align}
\Psi_{\mathrm{MSCN}} &: I \rightarrow \mathbf{f}_{\mathrm{MSCN}} \in \mathbb{R}^{d_1}, \\
\Psi_{\mathrm{CLIP}} &: I \rightarrow \mathbf{f}_{\mathrm{CLIP}} \in \mathbb{R}^{d_2}, \\
\Psi_{\mathrm{MLBP}} &: I \rightarrow \mathbf{f}_{\mathrm{MLBP}} \in \mathbb{R}^{d_3},
\end{align}
where $d_1, d_2, d_3$ are the dimensionalities of the respective feature vectors. The fused feature vector is then:
\begin{equation}
\mathbf{f}_{\mathrm{fused}} = [\mathbf{f}_{\mathrm{MSCN}} \ || \ \mathbf{f}_{\mathrm{CLIP}} \ || \ \mathbf{f}_{\mathrm{MLBP}}] \in \mathbb{R}^{d_1 + d_2 + d_3},
\end{equation}
where $||$ denotes concatenation.

\vspace{-0.1cm}
\subsection{Feature Space Extraction}

\subsubsection{MSCN Features: Capturing low-level statistical deviations}
Natural images exhibit well-defined statistical regularities in their luminance and contrast distributions. The mean subtracted contrast normalized (MSCN) transformation \cite{6272356} enhances these regularities for analysis. Given a grayscale image $I_{\text{gray}}$, the MSCN coefficients $C(i,j)$ are computed as:
\begin{equation}
C(i,j) = \frac{I_{\text{gray}}(i,j) - \mu(i,j)}{\sigma(i,j) + 1},
\end{equation}
where $\mu(i,j)$ and $\sigma(i,j)$ denote the local mean and standard deviation estimated using a $7 \times 7$ Gaussian window.

To characterize statistical deviation, we compute the gray-level co-occurrence matrix (GLCM) from the MSCN coefficients. The GLCM $P_{\Delta x, \Delta y}$ represents the joint probability of pixel pairs with values $i$ and $j$ at a given spatial displacement $(\Delta x, \Delta y)$. We consider four orientations ($0^\circ, 45^\circ, 90^\circ, 135^\circ$) with a pixel distance distance of 1. From each GLCM, we extract six Haralick features capturing complementary statistical properties:
\begin{itemize}
    \item \textit{Contrast}: Measures local intensity variations.
    \item \textit{Dissimilarity}: Similar to contrast but with linear increase.
    \item \textit{Homogeneity}: Measures closeness of element distribution to the diagonal.
    \item \textit{Angular second Moment (ASM)}: Measures uniformity.
    \item \textit{Energy}: Square root of ASM.
    \item \textit{Correlation}: Measures linear dependency of gray levels.
\end{itemize}
The final MSCN feature vector $\mathbf{f}_{\mathrm{MSCN}} \in \mathbb{R}^{72}$ is obtained by concatenating these features across all orientations.

\subsubsection{CLIP Embeddings: Encoding high-level semantic coherence}
Vision-Language Models (VLMs) such as CLIP \cite{pmlr-v139-radford21a} learn a joint embedding space where semantically related images and text are closely aligned. We leverage the pre-trained CLIP-ViT/B-32 model to extract high-level semantic representations. The image encoder $\mathrm{Encoder}_{\mathrm{CLIP}}$ processes the input image $I$ and outputs a normalized 512-dimensional vector:
\begin{equation}
\mathbf{f}_{\mathrm{CLIP}} = \mathrm{Encoder}_{\mathrm{CLIP}}(I) \in \mathbb{R}^{512}.
\end{equation}
This representation captures global scene semantics, object relationships, and contextual information. Even visually convincing GenAI images may contain subtle semantic inconsistencies, such as implausible interactions or physically unrealistic configurations, which can be reflected as deviations in this embedding space.

\subsubsection{MLBP Features: Characterizing mid-level texture anomalies}
Local binary patterns (LBP) \cite{pietikainen2010local} are effective texture descriptors that encode local spatial structures. For a grayscale image $I_{\text{gray}}$, the LBP operator at pixel $(x_c, y_c)$ with a neighborhood defined by radius $R$ and $P$ sampling points is defined as:
\begin{equation}
\mathrm{LBP}_{P,R}(x_c, y_c) = \sum_{p=0}^{P-1} s(g_p - g_c) 2^p, \quad s(x) = \begin{cases} 1, & x \geq 0 \\ 0, & x < 0 \end{cases}
\end{equation}
where $g_c$ and $g_p$ are the gray-level values of the center and neighboring pixels, respectively.

To capture multi-scale texture information, we employ a multi-scale LBP (MLBP) with radius $R \in \{1, 2, 3\}$. For each scale, we compute uniform LBP histograms (restricting patterns to at most two transitions), improving robustness while reducing dimentopinality. From each histogram $H_k$ (for scale $k$), we extract four statistical descriptions:
\begin{align}
\text{Mean:} & \quad \mu_k = \frac{1}{N} \sum_{i=0}^{L-1} i \cdot H_k(i), \\
\text{Variance:} & \quad \sigma_k^2 = \frac{1}{N} \sum_{i=0}^{L-1} (i - \mu_k)^2 \cdot H_k(i), \\
\text{Entropy:} & \quad E_k = -\sum_{i=0}^{L-1} \frac{H_k(i)}{N} \log\left(\frac{H_k(i)}{N} + \epsilon\right), \\
\text{Energy:} & \quad \mathrm{En}_k = \sum_{i=0}^{L-1} \left(\frac{H_k(i)}{N}\right)^2,
\end{align}
where $N$ is the total number of pixels, $L$ is the number of histogram bins, and $\epsilon$ is a small constant for numerical stability.

The final feature vector $\mathbf{f}_{\mathrm{MLBP}} \in \mathbb{R}^{36}$ is obtained by concatenating these descriptors across all scales. MLBP features are particularly sensitive to repetitive micro-patterns and boundary artifacts commonly introduced by generative models.

\begin{figure}[htbp]
    \centering
    \includegraphics[width=\linewidth]{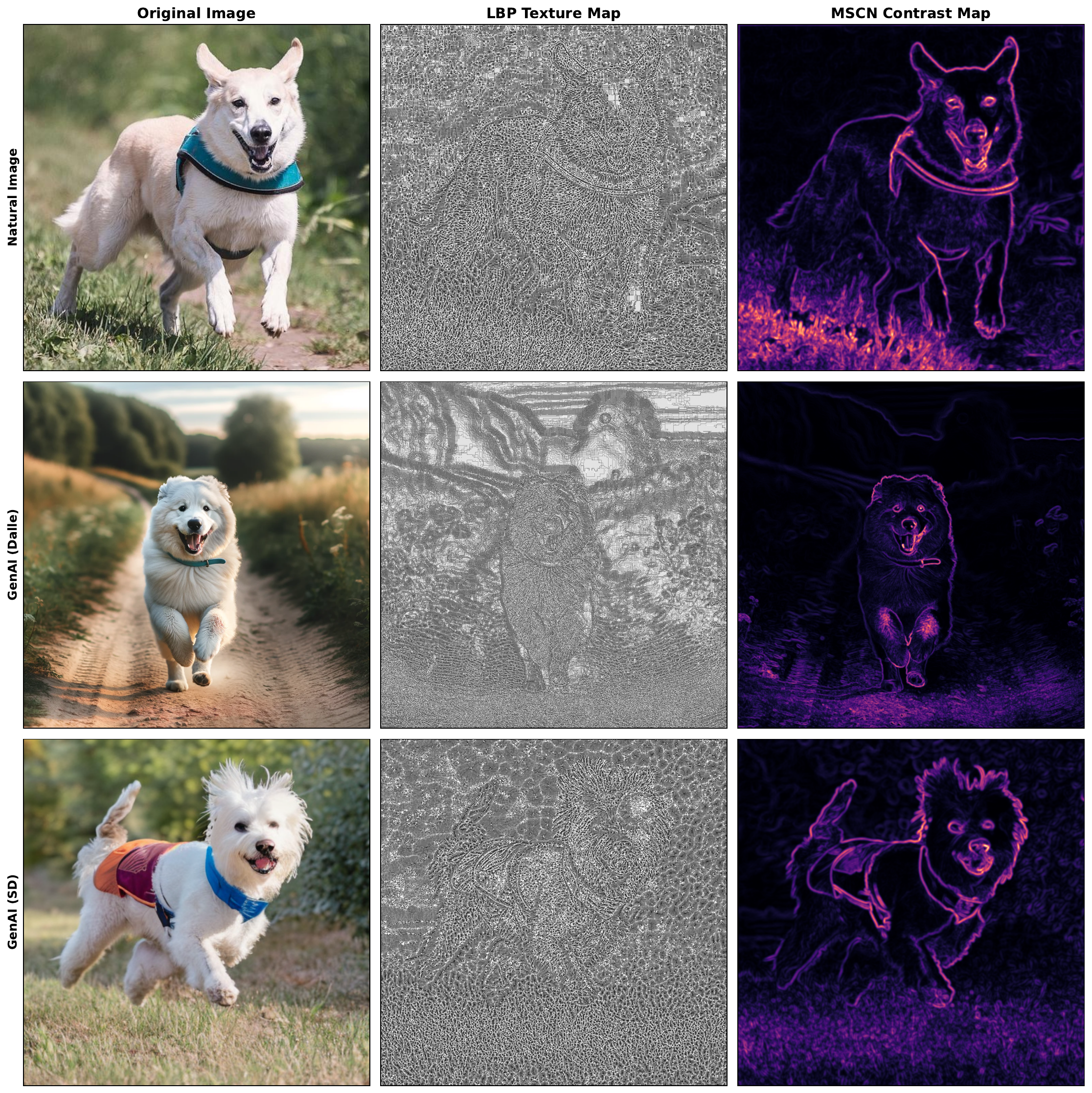}
    \caption{Visual comparison of texture (LBP) and contrast (GLCM) maps for a natural image and its AI-generated counterparts by DALLE and Stable Diffusion. The MSCN Contrast Maps (right) reveal that AI-generated images often exhibit inconsistent local intensity distributions and 'halo' artifacts compared to the smooth, natural scene statistics of the photograph. Simultaneously, the MLBP texture maps (center) expose the underlying structural regularity and 'tiling' artifacts inherent in generative architectures, contrasting sharply with the stochastic, high-variance micro-textures observed in the natural sample.}
    \label{fig:dist_maps}
\end{figure}

Figure \ref{fig:dist_maps} visualizes the LBP texture maps and GLCM contrast maps for a natural image and two GenAI examples, highlighting observable differences in texture regularity and contrast distribution that these feature spaces are designed to capture.

\vspace{-0.1cm}
\subsection{Feature Fusion and Classification}

\subsubsection{Normalization and Fusion}
To address the heterogeneous dimensionalities and dynamic ranges of the extracted feature sets, spanning the low-level statistical signatures, mid-level textural features, and high-level semantic embeddings, we propose a structured fusion function.

Prior to integration, each feature vector is standardized to a zero-mean, unit-variance space to prevent high-dimensional representations (\textit{e.g.}  CLIP embeddings) from dominating lower-dimensional features during optimization. Let $\mathbf{f}_k \in \{ \mathbf{f}_{\mathrm{MSCN}}, \mathbf{f}_{\mathrm{CLIP}}, \mathbf{f}_{\mathrm{MLBP}} \}$ denotes a raw feature vector. The normalized representation $\mathbf{z}_k$ is obtained as:

\begin{equation}
\mathbf{z}_k = \Gamma(\mathbf{f}_k) = \mathbf{V}_k^{-\frac{1}{2}} (\mathbf{f}_k - \mathbb{E}[\mathbf{f}k]),
\end{equation}
where $\mathbb{E}[\mathbf{f}_k]$ and $\mathbf{V}_k$ denote the expected mean and variance vectors estimated over the training manifold $\mathcal{D}_{\text{train}}$.

Rather than a simple stochastic stacking, we define the final Composite Image Fingerprint $\mathbf{F}$ through a fusion operator $\Phi$ that preserves the orthogonal properties of the three feature domains:
\begin{equation}
\mathbf{F} = \Phi(\mathbf{z}{\mathrm{MSCN}}, \mathbf{z}{\mathrm{CLIP}}, \mathbf{z}{\mathrm{MLBP}}) = \bigoplus{i=1}^{3} \mathbf{z}_i,
\end{equation}
where $\bigoplus$ represents the structured concatenation into a unified 620-dimensional feature space. This strategic aggregation ensures that the classifier $\mathcal{H}$ can learn inter-domain correlations, such as when a high-level semantic consistency (captured by CLIP) is contradicted by low-level statistical irregularities (captured by MSCN), a common trait in high-fidelity generative fakes.

The proposed fusion strategy is grounded in the multi-level nature of visual perception and the distinct failure modes of generative models:

\begin{itemize}
    \item \textbf{Statistical level (MSCN)}: Generative models, particularly earlier architectures or those trained on limited data, may deviate from natural luminance and contrast statistics. MSCN features directly capture these low-level anomalies.
    
    \item \textbf{Semantic level (CLIP)}: Even when low-level statistics are reproduced, generative models may exhibit semantic inconsistencies, such as objects with incorrect physical properties or inconsistent lighting directions. CLIP embeddings capture these high-level semantic discrepancies.
    
    \item \textbf{Textural level (MLBP)}: Generative models often struggle to reproduce the complex, non-repetitive textures found in natural images. MLBP features are sensitive to overly regular or artificial-looking textures that may appear in synthetic images.
\end{itemize}

By combining these three complementary levels, the proposed framework constructs comprehensive representations of image authenticity. A generative model that successfully fools one level of analysis (produces correct statistics) is unlikely to simultaneously fool all three. This defense-in-depth strategy is analogous to cryptographic security principles and provides robustness against the diverse and evolving outputs of modern GenAI models.

\subsubsection{Classification}
The final stage of the proposed framework maps the fused feature vector $\mathbf{F}$ to a probability space $\mathcal{P} \in [0, 1]$, where values close to 1 indicate a higher likelihood of generative origin. To identify the most effective decision boundary in this multi-modal feature space, we evaluate three structurally distinct learning paradigms:
\begin{itemize}
    \item Gradient boosting (GB): An ensemble technique that iteratively constructs a strong predictive model by minimizing a loss function through weak learners (typically decision trees). GB is well-suited for high-dimensional tabular data and offers implicit feature selection capabilities.
    \item Random forest (RF): A bagging-based ensemble method that aggregates multiple decorrelated decision trees. RF is included for its robustness to noise and its ability to mitigate overfitting.
    \item Support vector machines (SVM): A kernel-based method that seeks an optimal hyperplane maximizing class separation in the feature space. We utilize a radial basis function (RBF) kernel to capture non-linear relationships between the different feature domains.
\end{itemize}

Formally, for each classifier $\mathcal{H}$, the decision function is defined as:
\begin{equation}
\hat{y} = \mathbb{I}(\mathcal{H}(\mathbf{F}) > \tau)
\end{equation}
where $\mathbb{I}$ is the indicator function and $\tau$ is the decision threshold (typically set to 0.5). In Section \ref{sec:exp}, we provide a comparative evaluation of these classifiers to determine which model best leverages the complementarity of MSCN, CLIP, and MLBP features.

\vspace{-0.1cm}
\section{Experimental Setup and Results}
\label{sec:exp}

\subsection{Dataset}
To comprehensively evaluate the proposed framework across a wide range of generative models and architectures, we consider four benchmark datasets:

% \begin{description}
\textbf{Synthbuster \cite{10334046}}: this dataset contains 10k images collected to evaluate detection methods on content generated by modern GenAI models, including Stable Diffusion (versions 1.3, 1.4, 2, and XL), Midjourney, Adobe Firefly, and DALL·E (versions 2 and 3). Synthetic images are generated from text prompts inspired by the RAISE-1k dataset~\cite{dang2015raise}, which provides a collection of 1k high-quality, uncompressed images across diverse categories  (\textit{e.e.} landscapes, nature, objects, and buildings). Prompts are created using tools such as Midjourney descriptor \footnote{\href{https://docs.midjourney.com/hc/en-us/articles/32497889043981-Describe}{https://docs.midjourney.com/hc/en-us}} and CLIP Interrogator\footnote{\href{https://github.com/pharmapsychotic/clip-interrogator}{https://github.com/pharmapsychotic/clip-interrogator}}, and then manually refined to generate photorealistic images aligned with the original categories while preserving diversity.
        
\textbf{PKU-4K \cite{Yuan_2025_ICCV}}: this dataset includes approximately 4,000 AI-generated images covering both Text-to-Image (T2I) and Image-to-Image (I2I) generation paradigms. Images are produced using modern AI models, including Midjourney, Stable Diffusion v1.5, and DALL-E 3, with DALL-E 3. The latter is used exclusively for T2I tasks. The generation process relies on 200 diverse ImageNet-based prompts. High-resolution images from Pixabay are used as inputs for I2I generation, while textual descriptions guide T2I synthesis. A key characteristic of this dataset is the coexistence of both prompt modalities, enabling evaluation across heterogeneous generation settings.

\textbf{CIFAKE \cite{10409290}}: this dataset is a large-scale benchmark for synthetic vs. natural image composed of 120k RGB images: 60k natural drawn from the CIFAR-10~\cite{krizhevsky2009learning} dataset and 60k synthetic images generated by Stable Diffusion v1.4 trained on a subset of the LAION-5B corpus. The dataset spans ten object classes (airplane, automobile, bird, cat, etc.) at a resolution of $32\times32$, making it suitable for evaluating detection methods under constrained visual complexity and modest resolution scenarios. It emphasizes class-agnostic detection through binary (real/fake) labeling.

\textbf{FakeBench \cite{11124461}}: this dataset is a multimodal benchmark designed to evaluate both detection accuracy and interpretability. In this work, we focus only on the FakeClass subset, which contains 6k images (3k real and 3k AI-generated). Synthetic images are generated using a diverse set of models, including GAN-based approaches (ProGAN and the StyleGAN family) and diffusion-based models (DALL·E and Midjourney), ensuring coverage of heterogeneous synthesis artifacts. Real images are collected from multiple sources to ensure. Notably, the two classes do not necessarily share the same semantic content, making this dataset particularly relevant for evaluating robustness to distribution shifts and cross-domain generalization.
    
\vspace{-0.1cm}
\subsection{Evaluation Criteria}

To evaluate the performance of the proposed framework, we adopt Accuracy and the Matthews correlation coefficient (MCC) as primary metrics~\cite{Ruppert01062004, baldi2000assessing}. This choice ensures a balanced and informative assessment, especially in scenarios where class distributions may be imbalanced. Accuracy provides an intuitive measure of overall correctness but can be misleading in skewed datasets. To address this limitation, we complement it with MCC, a more robust metric that accounts for true and false positives and negatives, yielding a high score only when the classifier performs well across all components of the confusion matrix. MCC is widely regarded as a balanced measure in the presence of class size imbalance, making it particularly suitable for evaluating generalization under varying data distributions. Together, Accuracy, (ranging in $[0, 1]$ offers straightforward interpretability, while MCC (ranging in $[-1, 1]$ provides a more reliable and comprehensive evaluation of predictive performances across both classes.

\vspace{-0.1cm}
\subsection{Performance Evaluation}

\subsubsection{Individual vs. Fused features}

\begin{table}[htbp]
\centering
\renewcommand{\arraystretch}{1.1}
\setlength{\tabcolsep}{1.5pt}
\footnotesize
\caption{Performance comparison across the generative models in the Synthbuster dataset~\cite{10334046}.}
\label{tab:synthbuster_perfs}
\begin{tabular}{lcccccccccc}
\toprule
\textbf{Features} & \textbf{Met.} & \textbf{Dl-2} & \textbf{Dl-3} & \textbf{FF} & \textbf{Gld} & \textbf{MJ-5} & \textbf{S1.3} & \textbf{S1.4} & \textbf{S-XL} & \textbf{Avg.} \\ \midrule
\multirow{2}{*}{CLIP} & Acc. & 0.836 & 0.996 & 0.930 & 0.973 & 0.930 & 0.939 & 0.937 & 0.948 & 0.936 \\
 & MCC & 0.673 & 0.991 & 0.861 & 0.947 & 0.861 & 0.879 & 0.874 & 0.896 & 0.873 \\ \cmidrule{2-11}
\multirow{2}{*}{MLBP} & Acc. & 0.948 & 0.784 & 0.807 & 0.895 & 0.911 & 0.856 & 0.867 & 0.921 & 0.874 \\
 & MCC & 0.900 & 0.569 & 0.614 & 0.791 & 0.823 & 0.714 & 0.736 & 0.842 & 0.749 \\ \cmidrule{2-11}
\multirow{2}{*}{MSCN} & Acc. & 0.905 & 0.948 & 0.920 & 0.892 & 0.900 & 0.963 & 0.958 & 0.910 & 0.925 \\
 & MCC & 0.811 & 0.896 & 0.840 & 0.784 & 0.805 & 0.927 & 0.917 & 0.821 & 0.850 \\ \midrule\midrule
CLIP+ & Acc. & 0.929 & 0.995 & 0.941 & 0.978 & 0.941 & 0.955 & 0.955 & 0.958 & 0.957 \\
 MLBP & MCC & 0.861 & 0.989 & 0.882 & 0.956 & 0.882 & 0.911 & 0.911 & 0.916 & 0.915 \\ \cmidrule{2-11}
CLIP+ & Acc. & 0.900 & 0.993 & 0.947 & 0.979 & 0.956 & 0.971 & 0.970 & 0.958 & 0.959 \\
MSCN & MCC & 0.801 & 0.986 & 0.895 & 0.959 & 0.912 & 0.942 & 0.940 & 0.917 & 0.919 \\ \cmidrule{2-11}
MLBP+ & Acc. & 0.946 & 0.960 & 0.940 & 0.931 & 0.926 & 0.967 & 0.962 & 0.944 & 0.947 \\
MSCN & MCC & 0.894 & 0.921 & 0.880 & 0.864 & 0.855 & 0.935 & 0.925 & 0.889 & 0.895 \\ \cmidrule{2-11}
\multirow{2}{*}{ALL} & Acc. & 0.939 & 0.994 & 0.953 & 0.980 & 0.956 & 0.974 & 0.974 & 0.959 & 0.966 \\
 & MCC & 0.880 & 0.987 & 0.907 & 0.960 & 0.913 & 0.949 & 0.948 & 0.917 & 0.933 \\ \bottomrule
\end{tabular}
\end{table}

\begin{table}[htbp]
\centering
\renewcommand{\arraystretch}{1.1}
\setlength{\tabcolsep}{1.5pt}
\footnotesize
\caption{Performance comparison across the generative models in the PKU-4K dataset \cite{Yuan_2025_ICCV} under text-to-image (T2I) and image-to-image (I2I) scenarios.}
\label{tab:pku_perfs}
\begin{tabular}{l c ccccc cccc}
\toprule
\textbf{Features} & \textbf{Met.} & \multicolumn{5}{c}{\textbf{T2I}} & \multicolumn{3}{c}{\textbf{I2I}} \\ 
\cmidrule(lr){3-7} \cmidrule(lr){8-10}
 & & \textbf{MJ} & \textbf{SD} & \textbf{DALL·E} & \textbf{Avg.} & & \textbf{MJ} & \textbf{SD} & \textbf{Avg.} \\
\midrule
\multirow{2}{*}{CLIP}& Acc. & 0.9375 & 0.8411 & 0.9431 & 0.9072 & & 0.9181 & 0.8294 & 0.8738 \\
& MCC  & 0.8759 & 0.6911 & 0.8881 & 0.8184 & & 0.8424 & 0.6697 & 0.7561 \\\cmidrule{2-10}
\multirow{2}{*}{MSCN}& Acc. & 0.7615 & 0.7410 & 0.7331 & 0.7452 & & 0.7033 & 0.6619 & 0.6826 \\
& MCC  & 0.5300 & 0.4846 & 0.4719 & 0.4955 & & 0.4125 & 0.3329 & 0.3728 \\
\cmidrule{2-10}\multirow{2}{*}{MLBP}& Acc. & 0.8531 & 0.6281 & 0.6317 & 0.7043 & & 0.6750 & 0.5827 & 0.6289 \\
& MCC  & 0.7172 & 0.2597 & 0.2656 & 0.4142 & & 0.3540 & 0.1739 & 0.2639 \\
\midrule\midrule
CLIP +& Acc. & 0.9763 & 0.9100 & 0.9713 & 0.9525 & & 0.9394 & 0.8663 & 0.9029 \\
 MSCN& MCC  & 0.9532 & 0.8262 & 0.9440 & 0.9078 & & 0.8832 & 0.7421 & 0.8127 \\\cmidrule{2-10}
CLIP + & Acc. & 0.9825 & 0.9038 & 0.9781 & 0.9548 & & 0.9431 & 0.8819 & 0.9125 \\
MLBP& MCC  & 0.9652 & 0.8176 & 0.9569 & 0.9132 & & 0.8911 & 0.7726 & 0.8319 \\
\cmidrule{2-10}MLBP +& Acc. & 0.8625 & 0.8019 & 0.7669 & 0.8104 & & 0.7688 & 0.7225 & 0.7457 \\
 MSCN& MCC  & 0.7286 & 0.6065 & 0.5368 & 0.6240 & & 0.5395 & 0.4490 & 0.4943 \\
\cmidrule{2-10}\multirow{2}{*}{ALL}& Acc. & 0.9769 & 0.9088 & 0.9694 & 0.9517 & & 0.9400 & 0.8675 & 0.9038 \\
& MCC  & 0.9544 & 0.8233 & 0.9402 & 0.9060 & & 0.8844 & 0.7442 & 0.8143 \\
\bottomrule
\end{tabular}
\end{table}

\begin{table*}[htbp]
\centering
\renewcommand{\arraystretch}{1}
\setlength{\tabcolsep}{3pt}
\footnotesize
\caption{Performance comparison across generative models in the FakeBench dataset~\cite{11124461}.}
\label{tab:fakebench_perfs}
\begin{tabular}{lcccccccccccc}
\toprule
\textbf{Features} & \textbf{Met.} & \textbf{FD} & \textbf{Gld} & \textbf{VQDM} & \textbf{Cog2} & \textbf{Dl-2} & \textbf{Dl-3} & \textbf{MJ} & \textbf{ProG} & \textbf{SD} & \textbf{StyG} & \textbf{Avg.} \\
\midrule
\multirow{2}{*}{CLIP}
& Acc. & 0.973 & 0.955 & 0.842 & 0.955 & 0.948 & 0.927 & 0.968 & 0.936 & 0.884 & 0.903 & 0.929 \\
& MCC  & 0.946 & 0.913 & 0.691 & 0.914 & 0.898 & 0.868 & 0.937 & 0.877 & 0.786 & 0.832 & 0.866 \\
\cmidrule{2-13}
\multirow{2}{*}{MLBP}
& Acc. & 0.872 & 0.926 & 0.613 & 0.814 & 0.878 & 0.769 & 0.778 & 0.739 & 0.764 & 0.790 & 0.794 \\
& MCC  & 0.747 & 0.853 & 0.229 & 0.636 & 0.757 & 0.542 & 0.557 & 0.486 & 0.534 & 0.586 & 0.593 \\
\cmidrule{2-13}
\multirow{2}{*}{MSCN}
& Acc. & 0.784 & 0.921 & 0.632 & 0.796 & 0.854 & 0.784 & 0.804 & 0.751 & 0.720 & 0.808 & 0.785 \\
& MCC  & 0.570 & 0.844 & 0.267 & 0.595 & 0.712 & 0.571 & 0.609 & 0.503 & 0.443 & 0.625 & 0.574 \\
\midrule\midrule
\multirow{2}{*}{CLIP+MLBP}
& Acc. & 0.992 & 0.985 & 0.900 & 0.985 & 0.985 & 0.948 & 0.983 & 0.950 & 0.902 & 0.918 & 0.955 \\
& MCC  & 0.983 & 0.970 & 0.805 & 0.970 & 0.970 & 0.906 & 0.967 & 0.904 & 0.823 & 0.862 & 0.916 \\
\cmidrule{2-13}
\multirow{2}{*}{CLIP+MSCN}
& Acc. & 0.993 & 0.965 & 0.878 & 0.980 & 0.982 & 0.947 & 0.978 & 0.950 & 0.908 & 0.917 & 0.950 \\
& MCC  & 0.987 & 0.931 & 0.762 & 0.961 & 0.964 & 0.899 & 0.957 & 0.904 & 0.830 & 0.859 & 0.905 \\
\cmidrule{2-13}
\multirow{2}{*}{MLBP+MSCN}
& Acc. & 0.842 & 0.943 & 0.697 & 0.833 & 0.887 & 0.805 & 0.832 & 0.802 & 0.767 & 0.840 & 0.825 \\
& MCC  & 0.687 & 0.888 & 0.397 & 0.675 & 0.775 & 0.614 & 0.664 & 0.612 & 0.537 & 0.693 & 0.654 \\
\cmidrule{2-13}
\multirow{2}{*}{ALL}
& Acc. & 0.993 & 0.967 & 0.880 & 0.980 & 0.983 & 0.948 & 0.978 & 0.953 & 0.905 & 0.915 & 0.950 \\
& MCC  & 0.987 & 0.934 & 0.765 & 0.961 & 0.967 & 0.902 & 0.957 & 0.911 & 0.828 & 0.856 & 0.907 \\
\bottomrule
\end{tabular}
\end{table*}

To systematically evaluate the contribution of each feature space and their complementarity, we report Accuracy and MCC for individual representations (CLIP, MLBP, MSCN) as well as their fused variants (CLIP+MLBP, CLIP+MSCN, MLBP+MSCN, ALL) across the four benchmark datasets: Synthbuster (Table~\ref{tab:synthbuster_perfs}), PKU-4K (Table~\ref{tab:pku_perfs}), FakeBench (Table~\ref{tab:fakebench_perfs}), and CIFAKE (Table~\ref{tab:cifake_fakeb}). Results are reported per generator when available, along with average scores for global comparison.

\textbf{Synthbuster}: On Synthbuster, CLIP emerges as the strongest individual representation (Acc Avg. 0.936; MCC Avg. 0.873), confirming the effectiveness of high-level semantic embeddings for detecting fully synthetic images. MSCN follows closely (0.925 / 0.850), indicating that natural scene statistics (NSS) still capture meaningful structural deviations. In contrast, MLBP performs lower (0.874 / 0.749) and shows higher variability across generators.

Feature fusion consistently improves performance. Both CLIP+MSCN (0.959 / 0.919) and CLIP+MLBP (0.957 / 0.915) outperform their individual components, while ALL achieves the best overall results (0.966 / 0.933). Gains are systematic across generators and particularly pronounced in MCC, highlighting improved robustness and decision balance.

\textbf{PKU-4K (T2I vs. I2I)}: it introduces a more challenging setting, especially under I2I generation. For T2I, CLIP remains dominant (0.907 / 0.818), whereas MSCN (0.745 / 0.496) and MLBP (0.704 / 0.414) perform significantly worse. The gap widens in I2I, where all individual features degrade (CLIP: 0.874 / 0.756). This reflects the preservation of natural priors in I2I generation, where outputs remain anchored to real images, reducing detectable statistical and semantic inconsistencies. Consequently, I2I detection is substantially harder than T2I and more challenging than the fully synthetic Synthbuster setting.

\vspace{-0.3cm}
\begin{table}[htbp]
\centering
\renewcommand{\arraystretch}{1.1}
\setlength{\tabcolsep}{8pt}
\footnotesize
\caption{Performance comparison across generative models in the CIFAKE dataset~\cite{10409290}.}
\label{tab:cifake_fakeb}
\begin{tabular}{lcc}
\toprule
\textbf{Features} & \textbf{Acc.} & \textbf{MCC} \\ \midrule
MSCN          & 0.842 & 0.684 \\
MLBP          & 0.647 & 0.296 \\
CLIP          & 0.938 & 0.875 \\
\midrule \midrule
CLIP+MLBP     & 0.940 & 0.881 \\
CLIP+MSCN     & 0.944 & 0.888 \\
MLBP+MSCN     & 0.861 & 0.722 \\
ALL           & 0.946 & 0.892 \\ \bottomrule
\end{tabular}
\end{table}

In this regime, fusion becomes critical. CLIP+MLBP achieves the best performance (T2I: 0.955 / 0.913; I2I: 0.913 / 0.832), slightly outperforming CLIP+MSCN. Notably, the ALL configuration does not significantly exceed the best two-feature combinations, suggesting diminishing returns once complementary cues are effectively integrated.

\textbf{CIFAKE}: It provides a more controlled setting, enabling clearer analysis of feature behavior. CLIP again dominates among individual features (0.938 / 0.875), while MSCN (0.842 / 0.684) retains moderate discriminative power and MLBP drops significantly (0.647 / 0.296), confirming its limited standalone robustness.

Fusion consistently improves performances, with CLIP+MSCN (0.944 / 0.888) and CLIP+MLBP (0.940 / 0.881) outperform CLIP alone and ALL achieving the best results (0.946 / 0.892). Compared to PKU-4K and FakeBench, improvements are more moderate, suggesting that CIFAKE contains more explicit artifacts already captured by semantic features.

\textbf{FakeBench}: It introduces broader both GAN-based and diffusion-based generators, introducing broader synthesis diversity. CLIP remains the strongest individual baseline (0.929 / 0.866), while MLBP (0.794 / 0.593) and MSCN (0.785 / 0.574) lag behind. Certain generators (\textit{e.g.} VQDM), are consistently more challenging, reflecting improved realism.

Fusion substantially enhances robustness. CLIP+MLBP achieves the highest average MCC (0.916), followed by ALL (0.907) and CLIP+MSCN (0.905). Gains are particularly evident for diffusion-based models, highlighting the importance of combining semantic and statistical cues as generative realism improves.

\textbf{Cross-Dataset}: Several consistent patterns emerge across all four datasets:
\begin{itemize}
    \item Semantic dominance: CLIP is the strongest individual representation in every benchmark, indicating that semantic embeddings capture inconsistencies beyond low-level artifacts.
    \item Limit of single-feature approaches: MLBP and MSCN underperform when used alone, especially in high-realism scenarios (\textit{e.g.} PKU-4K I2I and FakeBench).
    \item Fusion synergy: Combining semantic and statistical features considerably improves MCC, confirming their complementarity.
    \item Impact of generation paradigm: Detection difficulty increases from fully synthetic settings (CIFAKE and Synthbuster) to heterogeneous realism (FakeBench) and peaks in I2I scenarios (PKU-4K), where natural image priors are preserved.
\end{itemize}

Overall, these results demonstrate that robust natural vs. GenAI discrimination requires the joint modeling of semantic coherence and deviations from natural image statistics. As generative models approach photorealistic fidelity, single-domain cues become insufficient, making multi-feature fusion essential for reliable detection.

\subsubsection{Manifold Divergence Interpretation}
\label{sec:manifold_interpretation}

The empirical behavior observed across Synthbuster, PKU-4K, FakeBench, and CIFAKE can be interpreted through a manifold perspective of natural image statistics. Natural images do not uniformly occupy the high-dimensional pixel space; instead, they lie on a highly structured manifold governed by physical image formation processes (\textit{e.g.} scene geometry, optics, illumination, and sensor characteristics) as well as intrinsic natural scene statistics. Generative models aim to approximate this manifold by learning the underlying data distribution. From this viewpoint, natural vs.\ GenAI detection can be framed as estimating the degree to which a generated sample adheres to or deviates from this manifold, where the notion of “closeness” inherently depends on the feature representation.

Different feature families probe complementary directions of this manifold. NSS-based descriptors such as MSCN capture low-order statistical regularities of natural scenes, including normalized luminance distributions and local spatial correlations \cite{ruderman1994statistics, mittal2012brisque}. When generative models fail to reproduce these properties, synthetic images exhibit measurable deviations in these subspaces. This explains the strong performance of NSS features on fully synthetic benchmarks (\textit{e.g. }Synthbuster, CIFAKE), where generated samples may diverge from the natural manifold along multiple low-level statistical dimensions. However, modern diffusion models \cite{ho2020ddpm, rombach2022ldm} explicitly learn denoising score functions and are therefore better equipped to match local statistics, increasing their alignment with the natural manifold in statistical projections. As a result, the discriminative power of purely NSS-based methods diminishes in more challenging scenarios, where synthetic samples become substantially indistinguishable from natural images in these low-level representations.

\begin{figure}[htbp]
\centering
\includegraphics[width=\linewidth]{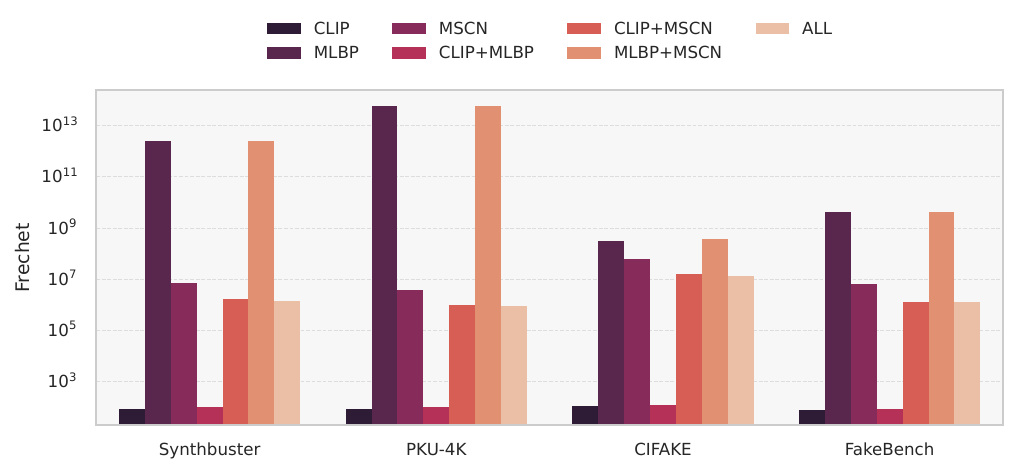}
\caption{Real--GenAI distribution separation across datasets and feature configurations. For each dataset, we report the (log-scaled) Gaussian Fréchet distance (FID-like) between the natural and GenAI feature distributions for the seven configurations (three individual feature families and their fusions). Larger values indicate greater deviation from the natural image manifold in the corresponding feature projection, while smaller values indicate stronger manifold overlap and a harder detection regime.}
\label{fig:frechet_distance}
\end{figure}

\begin{figure}[htbp]
\centering
\includegraphics[width=\linewidth]{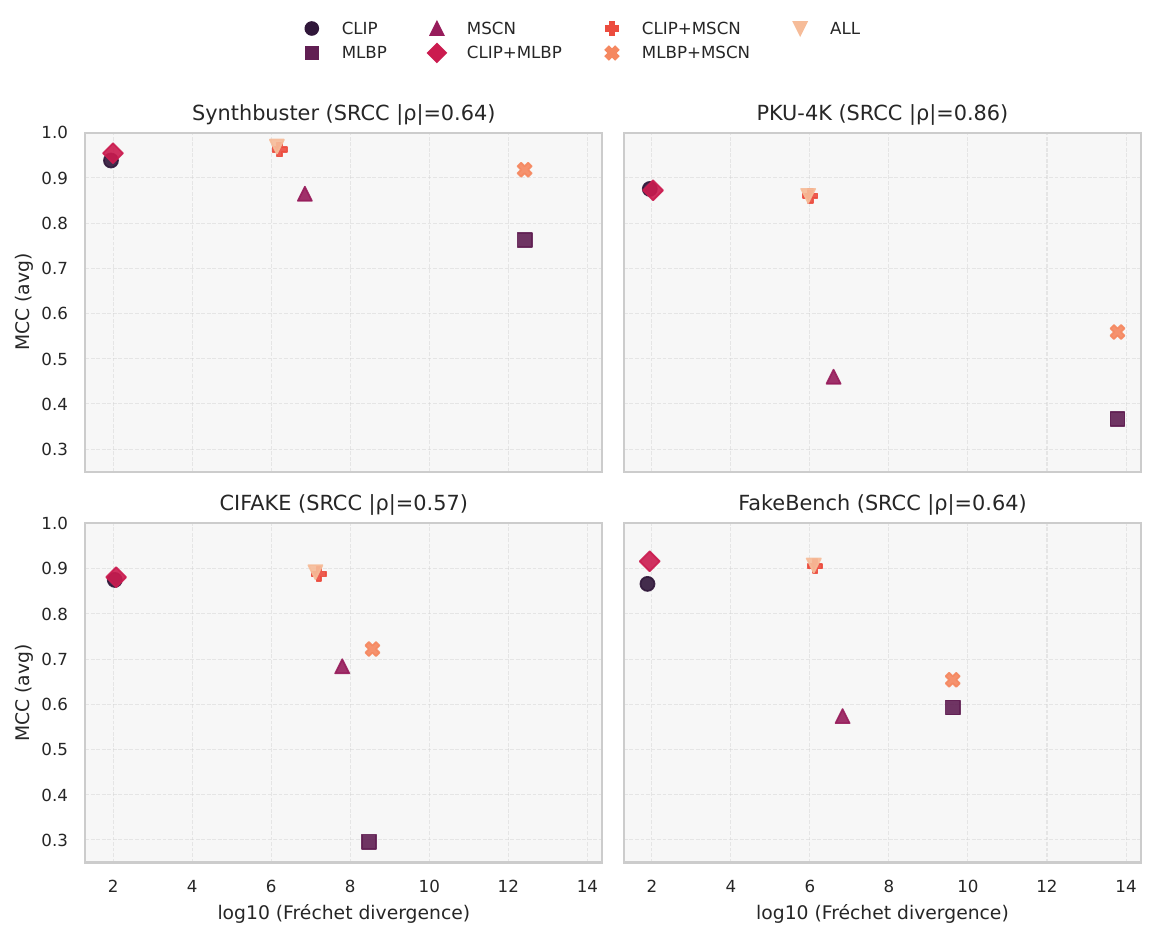}
\caption{Association between manifold separation and detection performance across feature configurations. For each dataset (Synthbuster, PKU, CIFAKE, FakeBench), points correspond to the seven feature configurations (marker-coded) and plot average MCC (y-axis) versus log10 Fréchet divergence between natural and GenAI feature distributions (x-axis). The title of each panel reports the absolute Spearman rank correlation (SRCC $|\rho|$) computed across configurations.}\vspace{-0.5cm}
\label{fig:correlation_frechet_mcc}
\end{figure}
% , summarizing the strength of the monotonic relationship between representation-level separation and detection performance within that dataset.

Semantic embeddings derived from CLIP \cite{radford2021clip} operate at a higher level of abstraction. Rather than capturing low-level statistical conformity, they encode semantic coherence, object relationships, and contextual consistency learned from large-scale vision-language supervision. In text-to-image (T2I) generation, subtle semantic inconsistencies or compositional irregularities can shift synthetic samples away from the natural manifold in this semantic space, explaining the consistent dominance of CLIP-based configurations across datasets. In contrast, image-to-image (I2I) generation starts from real images and largely preserves global semantics; as a result, generated samples remain closer to the natural manifold in semantic projections, consistent with the performance degradation observed in PKU-4K I2I.

The type of generative model further infleunces manifold alignment. Early GAN-based approaches such as ProGAN and StyleGAN \cite{karras2018progan, karras2019stylegan} may reproduce local texture statistics convincingly while introducing frequency artifacts or global structural inconsistencies. In contrast, diffusion models progressively refine noise toward data-consistent solutions, producing samples that more closely align with both statistical and semantic properties of natural images. Consequently, detection difficulty correlates with the degree of manifold overlap: fully synthetic benchmarks (\textit{e.g.} CIFAKE, Synthbuster) exhibit clearer separability, whereas image-conditioned settings regimes (\textit{e.g.} PKU-4K I2I) approach near-manifold alignment and represent significantly more challenging detection regimes.

To qualitatively support this manifold interpretation, we measure the separation between natural and GenAI samples in each feature space using a Gaussian Fréchet distance (FID-like) their distributions. Under this framework, a larger Fréchet distance indicates greater deviation from the natural manifold \emph{in the chosen representation}, wheras a smaller value reflect stronger overlap and increased detection difficulty. As shown in Figure~\ref{fig:frechet_distance} the magnitude of this separation varies significantly across both datasets (generation regime) and feature families (manifold projection).

We further relate separation to detection performances by analyzing the correlation between Fréchet distance and MCC across feature configurations (Figure~\ref{fig:correlation_frechet_mcc}). Within each dataset, greater distributional separation generally corresponds to higher MCC, yielding moderate-to-strong rank correlation (SRCC). This supports the claim that detectability is governed by the degree of manifold overlap in the selected representation space. However, Fréchet distance alone is not sufficient to predict performances across different feature families. Some low-level descriptors can exhibit large global distribution shifts while remaining weakly discriminative, whereas CLIP-based embeddings can achieve high MCC despite smaller Fréchet distances. This suggests that performance depends not only on the magnitude of distributional shift but also on its alignment with class-discriminative directions. 

From this perspective, feature fusion can be interpreted as multi-directional manifold probing. By combining complementary projections, statistical regularity (MSCN), local micro-structure (MLBP), and semantic consistency (CLIP), the proposed framework reduces blind spots arising from partial manifold alignment in any single subspace, thereby improving robustness and overall detection performance.

\subsubsection{Visualizing Feature Spaces}

To complement the quantitative analysis, we provide a qualitative embedding study based on t-SNE to visually assess the separability between natural and GenAI samples under different feature configurations. We focus on CIFAKE for two reasons. First, it involves a single generative source, eliminating the confounding effects from heterogeneous generators and allowing differences in embedding structure to be attributed primarily to the feature representation. Second, CIFAKE offers the largest sample size among the considered datasets, enabling a high-density visualization while ensuring fairness by using the same sampled images across all configurations. This setting provides a controlled and well-powered case study to illustrate the benefits of feature fusion in embedding class separability.

\begin{figure*}[htbp]
    \centering
    \includegraphics[width=\linewidth]{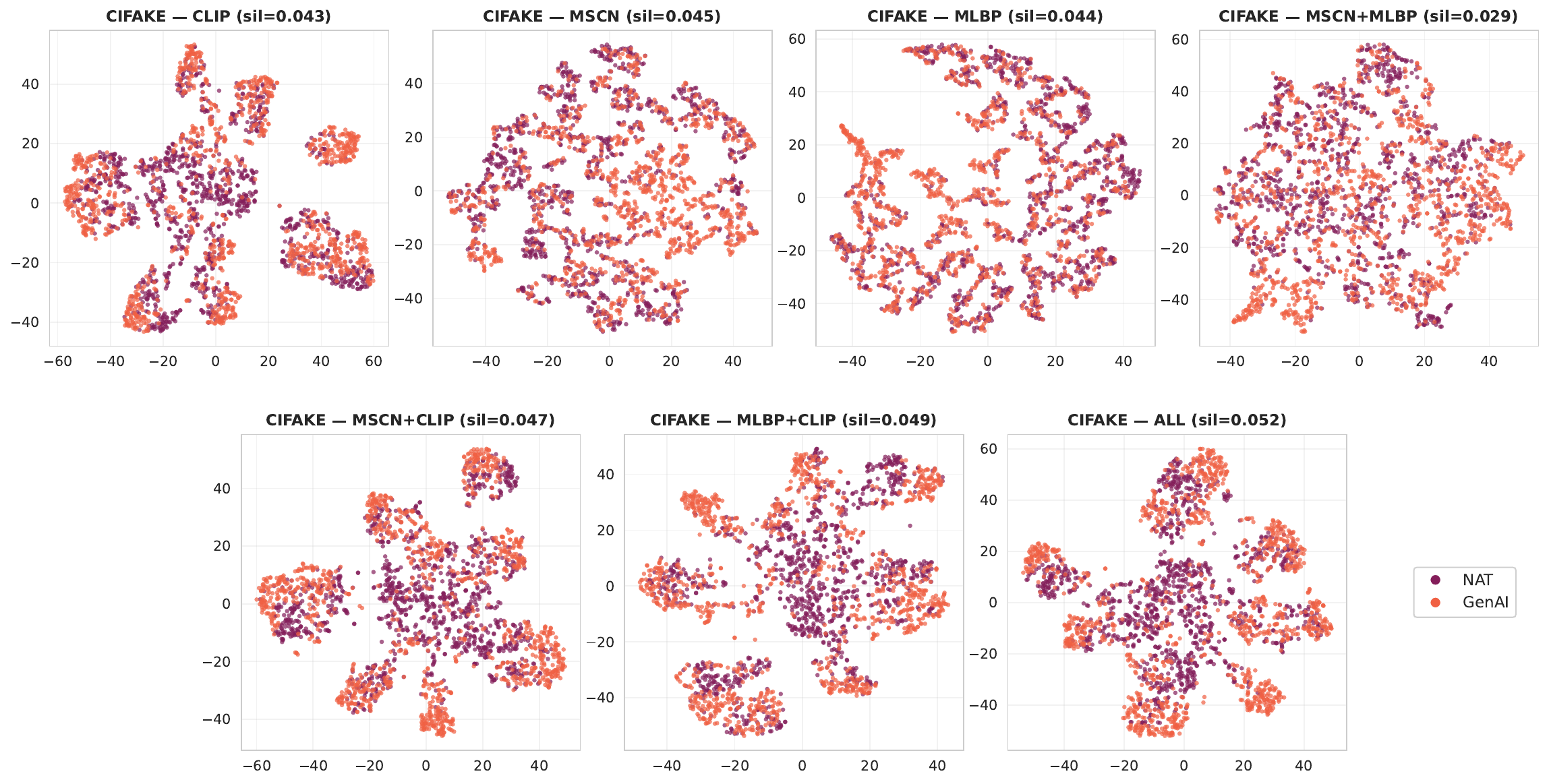}
    \caption{t-SNE on CIFAKE (natural vs. GenAI) across all feature configurations}\vspace{-0.5cm}
    \label{fig:tsne_cifake_fusion}
\end{figure*}

Figure~\ref{fig:tsne_cifake_fusion}  visualizes the separation between natural vs. GenAI samples in CIFAKE across seven feature configurations (three individual representations, three pairwise fusions, and full fusion). Individual feature spaces exhibit noticeable class overlap, whereas fused representations produce more compact and better separated clusters. This observation supports the hypothesis that semantic cues (CLIP) and low-level statistical texture descriptors (MSCN/MLBP) capture complementary information. To avoid over-interpreting a 2D t-SNE projection, we complement the visualization with an objective separability measure, namely the silhouette score, computed in the standardized and PCA-reduced space used prior to t-SNE. Under this controlled protocol, the full fusion \textbf{ALL} achieves the highest silhouette score ($0.052$), outperforming the best pairwise fusion (MSCN+CLIP: $0.047$), and clearly surpassing individual representations (\textit{e.g.} CLIP: $0.043$). Although absolute values remain modest, indicating that CIFAKE is not perfectly separable, the relative ranking confirms a consistent trend: combining complementary feature spaces yields the most discriminative representation.

\subsubsection{Statistical analysis}

\begin{figure*}[htbp]
    \centering
    \begin{tabular}{c}
         \includegraphics[width=0.9\linewidth]{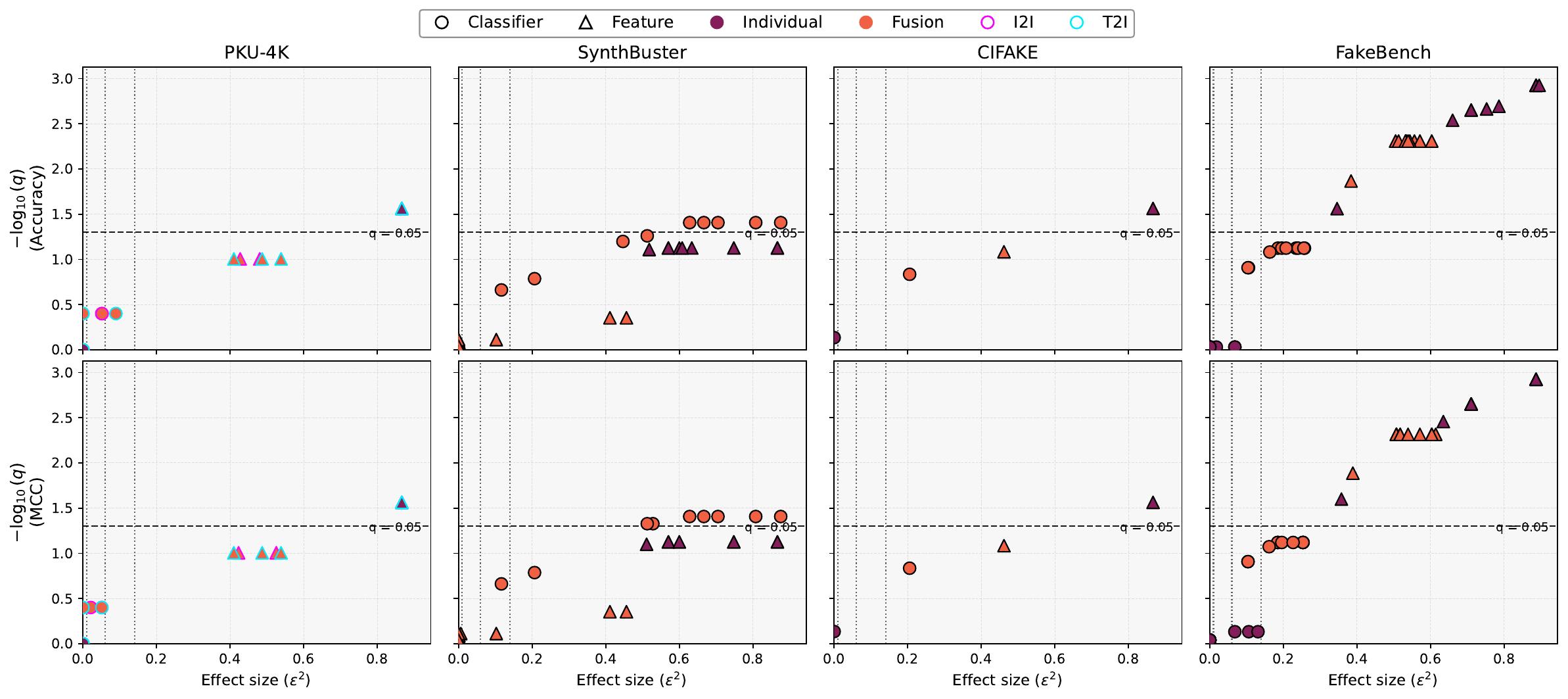} \\
    \end{tabular}
    \caption{Effect Size vs. Significance for classifier and feature effects by metrics (Accuracy: top and MCC: bottom). Each point corresponds to a generator-level Kruskal–Wallis test after FDR correction. Marker shape encodes the tested factor (circles: classifier; triangles: feature), and color encodes the feature categories. The x-axis reports effect size $\epsilon^2$ (0–1), the y-axis shows significance as $-\log_{10}(q)$. The dashed horizontal line marks the $q=0.05$ threshold; vertical dotted lines indicate small/medium/large effect-size heuristics. Labels are shown only for statistically significant points to reduce clutter.}\vspace{-0.5cm}
    \label{fig:significance_indiv_synth}
\end{figure*}

To quantify the relative impact of representation and classifier design, we conduct a generator-level statistical analysis across all datasets. For each dataset × generator × metric configuration, we apply the Kruskal–Wallis test to compare performance distributions induced by either feature choice or classifier selection. Effect sizes are reported using $\epsilon^2$, and statistical significance is assessed after Benjamini–Hochberg discovering rate (FDR) correction~\cite{benjamini1995controlling}. Both Accuracy and MCC are analyzed to ensure metric-independent conclusions. By jointly analyzing effect size and corrected significance, we distinguish statistically detectable differences from practically meaningful improvements. The result of this analysis is depicted in Figure~\ref{fig:significance_indiv_synth}.

% \begin{figure}[htbp]
%     \centering
%     \begin{tabular}{c}
%          (a) Synthbuster \\   
%          \includegraphics[width=.7\linewidth]{Figs_v1/effect_size_vs_significance_split_synthbuster.pdf} \\
%          (b) PKU-4K \\
%          \includegraphics[width=.7\linewidth]{Figs_v1/effect_size_vs_significance_split_pku.pdf} \\
%           (c) CIFAKE \\
%          \includegraphics[width=.7\linewidth]{Figs_v1/effect_size_vs_significance_split_cifake.pdf} \\
%           (d) FakeBench \\
%          \includegraphics[width=.7\linewidth]{Figs_v1/effect_size_vs_significance_split_fakebench.pdf} \\
%     \end{tabular}
%     \caption{Effect Size vs. Significance for classifier and feature effects by metrics (Accuracy and MCC). Each point corresponds to a generator-level Kruskal–Wallis test after FDR correction. Marker shape encodes the tested factor (circles: classifier; triangles: feature), and color encodes the feature categories. The x-axis reports effect size $\epsilon^2$ (0–1), the y-axis shows significance as $-\log_{10}(q)$. The dashed horizontal line marks the $q=0.05$ threshold; vertical dotted lines indicate small/medium/large effect-size heuristics. Labels are shown only for statistically significant points to reduce clutter.}
%     \label{fig:significance_indiv_synth}
% \end{figure}

On Synthbuster, a clear transition emerges between the Individual and Fusion configurations. When features are evaluated separately, performance variability is primarily driven by representation choice: feature effects are moderate and occasionally significant, while classifier effects remain weak and inconsistent. In this scenario, detection is representation-dominated. After fusion, the pattern shifts statistical classifier effects increase in magnitude and frequently reach significance, while feature effects diminish. Once complementary cues are integrated, the decision layer begins to shape the boundary in a statistically reliable manner. Synthbuster thus exhibits a two-stage behavior: representation-driven with individual features and classifier-influenced after fusion.

PKU-4K reveals a different pattern. Across both Individual and Fusion settings, classifier effects remain negligible, with effect sizes close to zero and rarely significant. In contrast, feature effects under the Individual setting are large and consistently significant, indicating that representation design fully governs performance. After fusion, these differences collapse: effect sizes decrease and significance disappears. Fusion homogenizes the representation space, acting as a variance stabilizer. PKU-4K therefore operates in a strongly feature-dominated regime, largely insensitive to classifier choice.

CIFAKE confirms this behavior in a more controlled binary setting. Under individual features, representation effects are moderate to large and often significant, while classifier effects remain negligible. After fusion, effect sizes contract and lose statistical significance, leading to stable performance across configurations. This reinforces the conclusion that, in balanced datasets with limited generator diversity, representation geometry is the primary driver of detection variability.

FakeBench introduces greater heterogeneity, spanning both GAN-based and diffusion-based generators. In the Individual setting, feature effects are again substantial and often significant. Under fusion, however, classifier effects do not fully vanish, and in some cases, approach or reach statistical relevance. This suggests that heterogeneous artifact distributions induce residual sensitivity to the decision layer once strong fused representations are available. FakeBench thus represents an intermediate scenario: feature-dominated initially, with measurable classifier interaction after fusion.

Across all datasets, three consistent principles emerge. First, in the Individual setting, representation choice dominates performance variability, with moderate-to-large and often significant effect sizes. Second, feature fusion systematically reduces variability, shrinking effect sizes and weakening statistical significance, acting as a stabilizing mechanism through the integration of complementary cues. Third, classifier influence is dataset-dependent and becomes relevant only when representation quality is high and residual heterogeneity remains.

These findings admit a geometric interpretation. Natural and generated images can be viewed as occupying partially overlapping regions on a high-dimensional visual manifold. Feature extractors define coordinate systems on this manifold, emphasizing different properties such as semantic consistency or statistical regularity. When considered separately, these representations induce distinct separability structures, explaining the strong feature effects observed in the Individual setting. Fusion embeds samples into a rich multidimensional space that captures complementary deviation from the natural manifold, reducing variability and stabilizing separability. Once this geometric structure is established, classifier influence diminishes unless residual heterogeneity persists, as observed in FakeBench and partially in Synthbuster.

Overall, results consistently indicate that natural versus GenAI detection is fundamentally a representation-driven problem. While classifiers refine decision boundaries, the primary determinant of performance lies in how effectively the feature space exposes deviations from the natural image manifold. Feature fusion reduces variance, enhances robustness, and shifts the problem from representation sensitivity toward geometric stabilization.

\begin{figure*}[htbp]
    \centering
    \begin{tabular}{c}
         \includegraphics[width=0.9\linewidth]{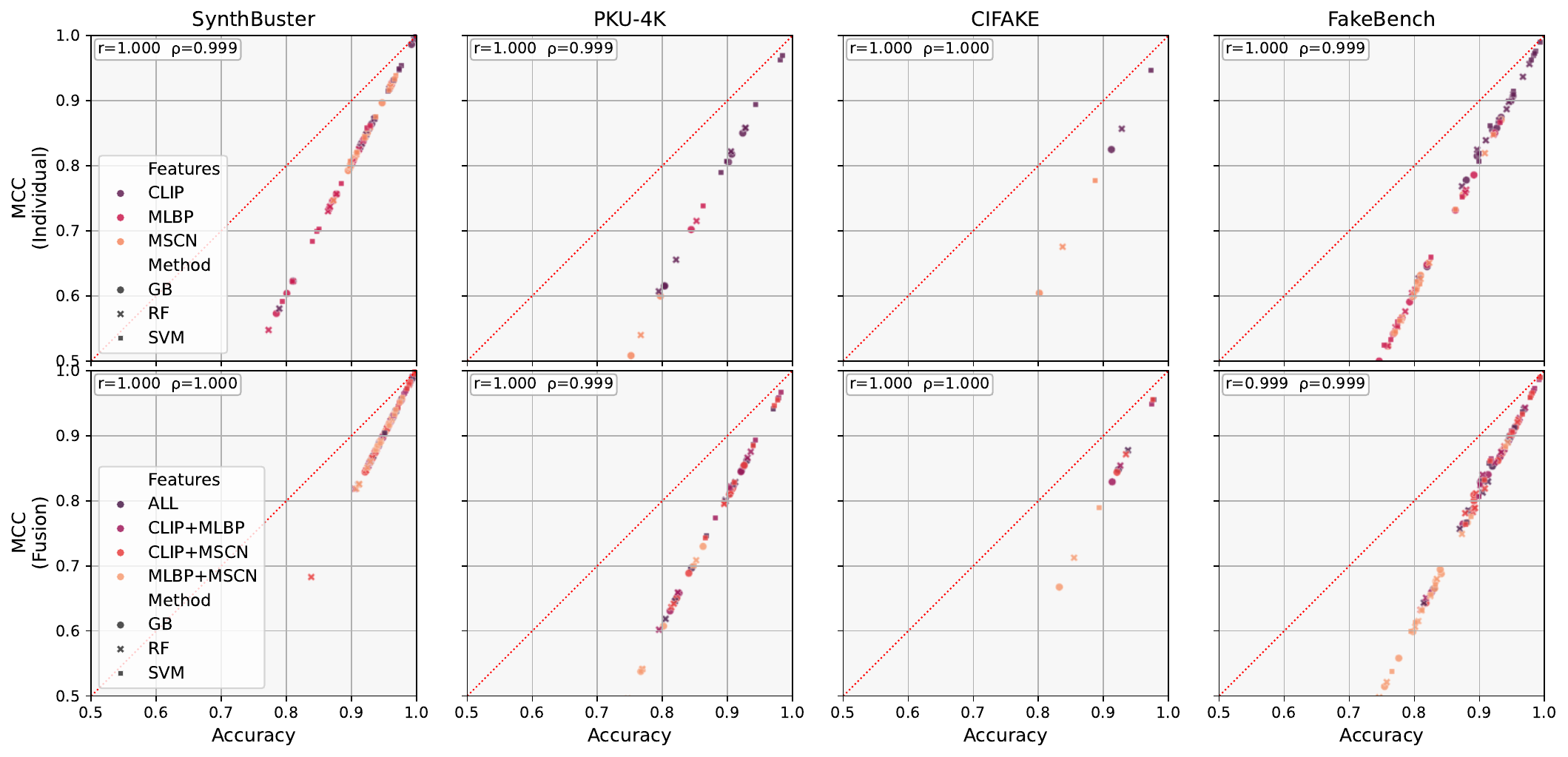}\\ 
    \end{tabular}
    \caption{Accuracy–MCC correlation for Individual (top) vs Fusion (bottom) features. The scatter points are colored by feature and styled by classifier.}\vspace{-0.5cm}
    \label{fig:acc_mcc}
\end{figure*}

% \begin{figure}[htbp]
%     \centering
%     \begin{tabular}{c}
%             (a) Synthbuster \\
%          \includegraphics[width=.7\linewidth]{Figs_v1/Acc_mcc_corr_synthbuster.pdf} \\
%             (b) PKU-4K \\
%          \includegraphics[width=.7\linewidth]{Figs_v1/Acc_mcc_corr_pku.pdf}\\ 
%           (c) CIFAKE \\
%          \includegraphics[width=.7\linewidth]{Figs_v1/Acc_mcc_corr_synthbuster.pdf} \\
%             (d) FakeBench \\
%          \includegraphics[width=.7\linewidth]{Figs_v1/Acc_mcc_corr_pku.pdf}\\ 
%     \end{tabular}
%     \caption{Accuracy–MCC correlation for Individual (left) vs Fusion (right) features. The scatter points are colored by feature and styled by classification method.}
%     \label{fig:acc_mcc}
% \end{figure}

Figure~\ref{fig:acc_mcc} presents the Accuracy–MCC scatter plots for individual and fused feature representations on Synthbuster and PKU-4K. While Accuracy and MCC are globally correlated, the dispersion and clustering patterns reveal important differences in reliability and decision balance.

For individual features, the scatter exhibits substantial spread, particularly along the MCC axis. MLBP and MSCN frequently achieve moderate accuracy with low MCC, indicating biased or unstable decision boundaries. CLIP forms a higher-performing cluster, yet still exhibits noticeable variability across classifiers, especially on PKU-4K, reflecting sensitivity to both model choice and dataset complexity. In contrast, representation features consistently form compact clusters in the upper-right region of the plots, with simultaneously high Accuracy and MCC. This concentration indicates improved calibration, stringer decision balance, and reduced dependence on the classifier. The effect is particularly pronounced on PKU-4K, where individual features are widely dispersed due to the realism of I2I generation, while fusion maintains stable and robust performance.

Across datasets, individual features behave more consistently on Synthbuster, where generative artifacts are more pronounced, but degrade significantly on PKU-4K. Fusion effectively mitigates this shift, preserving a strong Accuracy and MCC coupling under more challenging conditions.

Overall, the scatter analysis demonstrates that feature fusion improves not only peak performance but also robustness and decision stability, reinforcing the necessity to combine semantic, statistical, and textural cues for reliable GenAI image detection.

\begin{figure}[htbp]
    \centering
    \begin{tabular}{c}
        \includegraphics[width=\linewidth]{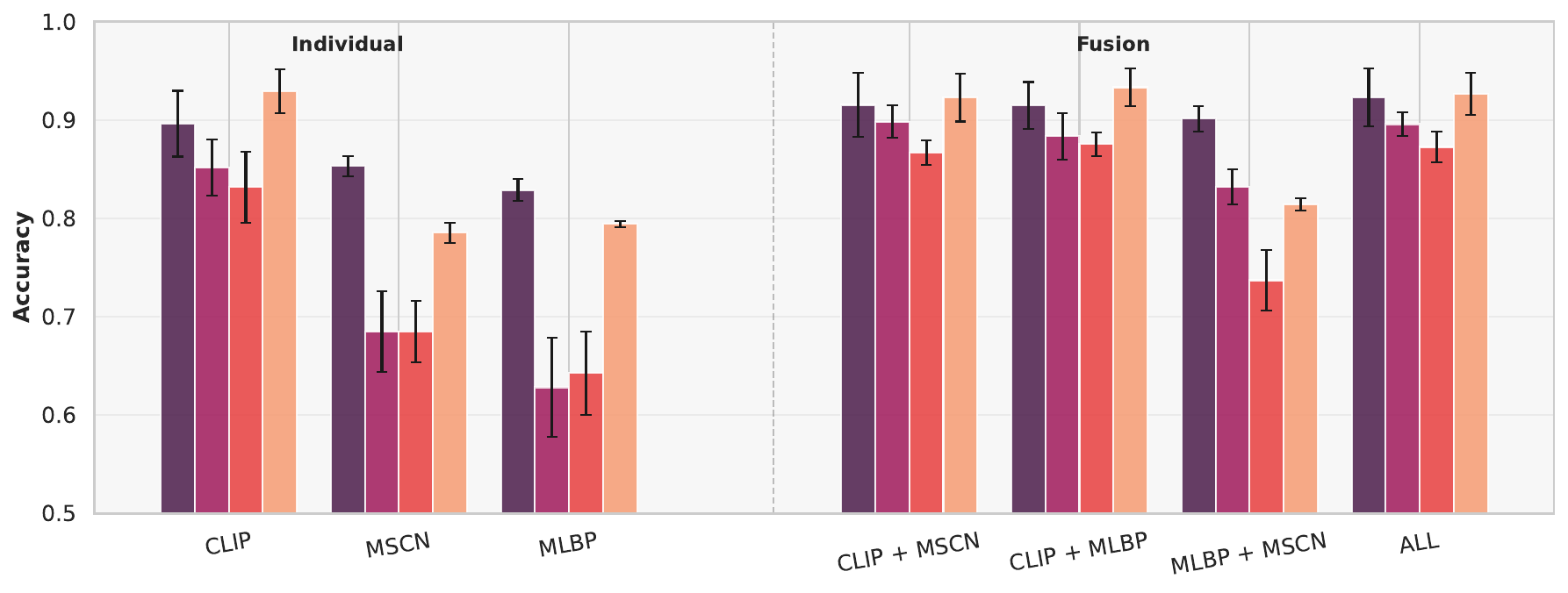}\\
        \includegraphics[width=\linewidth]{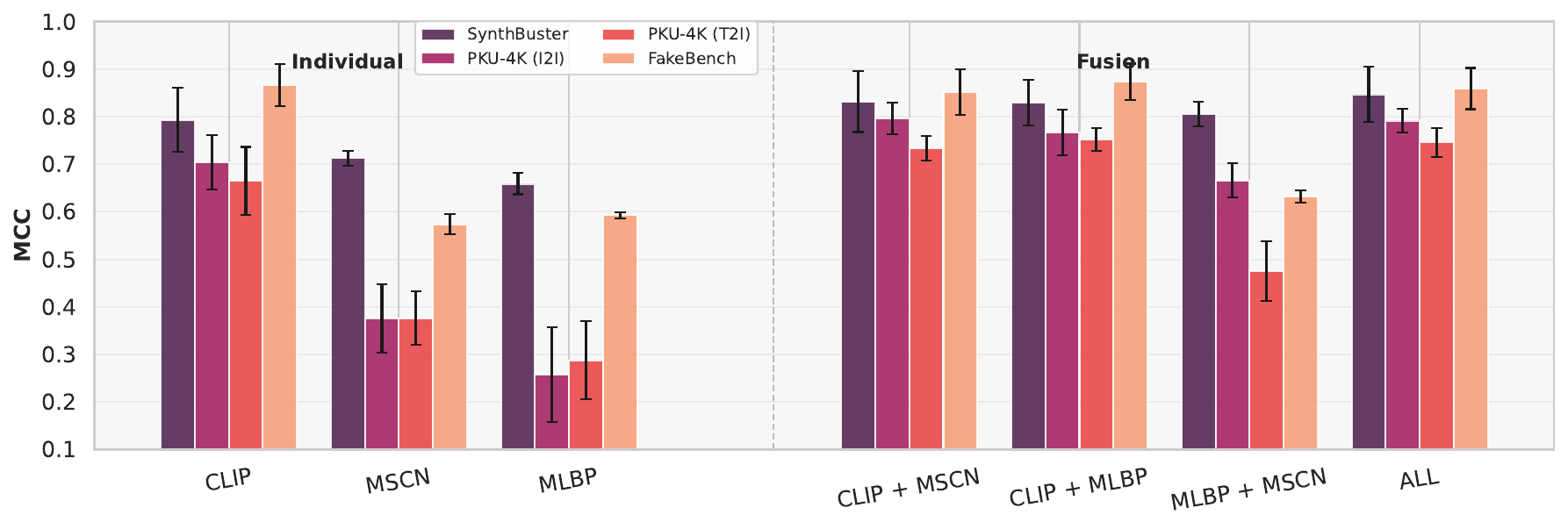}\\
    \end{tabular}
    \caption{Generalizability comparison of individual and fusion features across Synthbuster, PKU-4K (T2I/I2I), and FakeBench.}\vspace{-0.5cm}
    \label{fig:random_genai_bars}
\end{figure}

\vspace{-0.1cm}
\subsection{Robustness Under Randomized GenAI Composition}
\label{subsec:random_genai}

Beyond generator-specific discrimination, we evaluate the robustness of the proposed representations under a random GenAI setting, where the synthetic class is formed by randomly sampling images originating from multiple generative models. This protocol better reflects deployment scenarios, where detectors face heterogeneous and evolving generative sources. To reduce dependence on a specific learning backend, results are averaged across three standard classifiers (Gradient Boosting, Random Forest, and SVM), with bars indicating the mean Accuracy and MCC, and error bars representing standard deviation. Figure~\ref{fig:random_genai_bars}.

On Synthbuster, high-level semantics generalize well under heterogeneous generation. CLIP achieves the strongest individual performance (Acc. $0.896\pm0.033$, MCC $0.793\pm0.067$), while MSCN and MLBP degrade, particularly in MCC. Fusion yields systematic gains in both performance and stability, with ALL achieving the best results (Acc. $0.923\pm0.030$, MCC $0.847\pm0.059$). These improvements are accompanied by reduced variance across classifiers, suggesting more reliable decision boundaries.

PKU-4K presents a substantially more challenging setting. CLIP remains the strongest individual representation (I2I: Acc. $0.852\pm0.029$, MCC $0.704\pm0.057$; T2I: Acc. $0.832\pm0.036$, MCC $0.665\pm0.071$), whereas MSCN and especially MLBP exhibit severe degradation (MCC $\approx0.26$--$0.38$), indicating the limitation of handcrafted statistics in realistic generation regimes. Fusion recovers a large portion of this performance loss: in I2I, CLIP+MSCN and ALL approach $0.90$ Accuracy with MCC around $0.79$--$0.80$, while in T2I, fusion improves robustness Acc. $\approx0.87$ and MCC $\approx0.75$. Notably, I2I slightly outperforms T2I under fusion, suggesting that relative difficulty is data- and protocol-dependent, although both remain markedly harder than Synthbuster for low-level-only features.

On FakeBench, CLIP alone is already strong (Acc. $0.926\pm0.022$, MCC $0.850\pm0.044$), and fusion provides moderate but consistent gains. The best results remain CLIP-centered (\textit{e.g.} CLIP+MLBP $0.930\pm0.019$ / $0.862\pm0.038$), while ALL preserves strong performance ($0.939\pm0.022$ / $0.883\pm0.043$). In this setting, fusion primarily acts as a stabilizer rather than a fundamental performance driver.

Across datasets, three consistent observations emerge. First, semantic representations (CLIP) are essential for generalization under heterogeneous generator mixtures. Second, multi-feature fusion systematically improves both performance and stability, with the largest gains observed on PKU-4K. Third, low-level features alone are brittle, particularly visible in MCC on PKU-4K, whereas fusion mitigates this limitation by integrating complementary cues. Overall, CLIP-centered fusion provides a robust and deployment-oriented strategy for GenAI image detection in realistic conditions.

\vspace{-0.1cm}
\subsection{Performance Comparison with SOTA}

To demonstrate the effectiveness of the proposed framework, we compared it with several state-of-the-art detectors, including CNNDetection~\cite{wang2019cnngenerated} with two configurations (0.1 and 0.5), DMImageDetection~\cite{10095167} with two configurations (Latent and ProGAN), PatchForensics~\cite{patchforensics}, and Synthbuster~\cite{10334046}. These methods represent different paradigms. From CNN-based classifiers to patch-level forensic analysis and diffusion-specific frequency methods. CNNDetection relies on supervised convolutional classifiers trained to discriminate real from GAN-generated images, while DMImageDetection specifically targets diffusion-based artifacts through feature analysis in both latent and image domains. PatchForensics focuses on local forensic traces by analyzing patch-level inconsistencies that emerge from generative synthesis processes. Synthbuster is a frequency analysis based method that uese cross-difference filters, designed for diffusion-generated images.

% detectors, we aim to provide a broad and representative evaluation of current detection capabilities.

\begin{figure}[htbp]
    \centering
    \includegraphics[width=\linewidth]{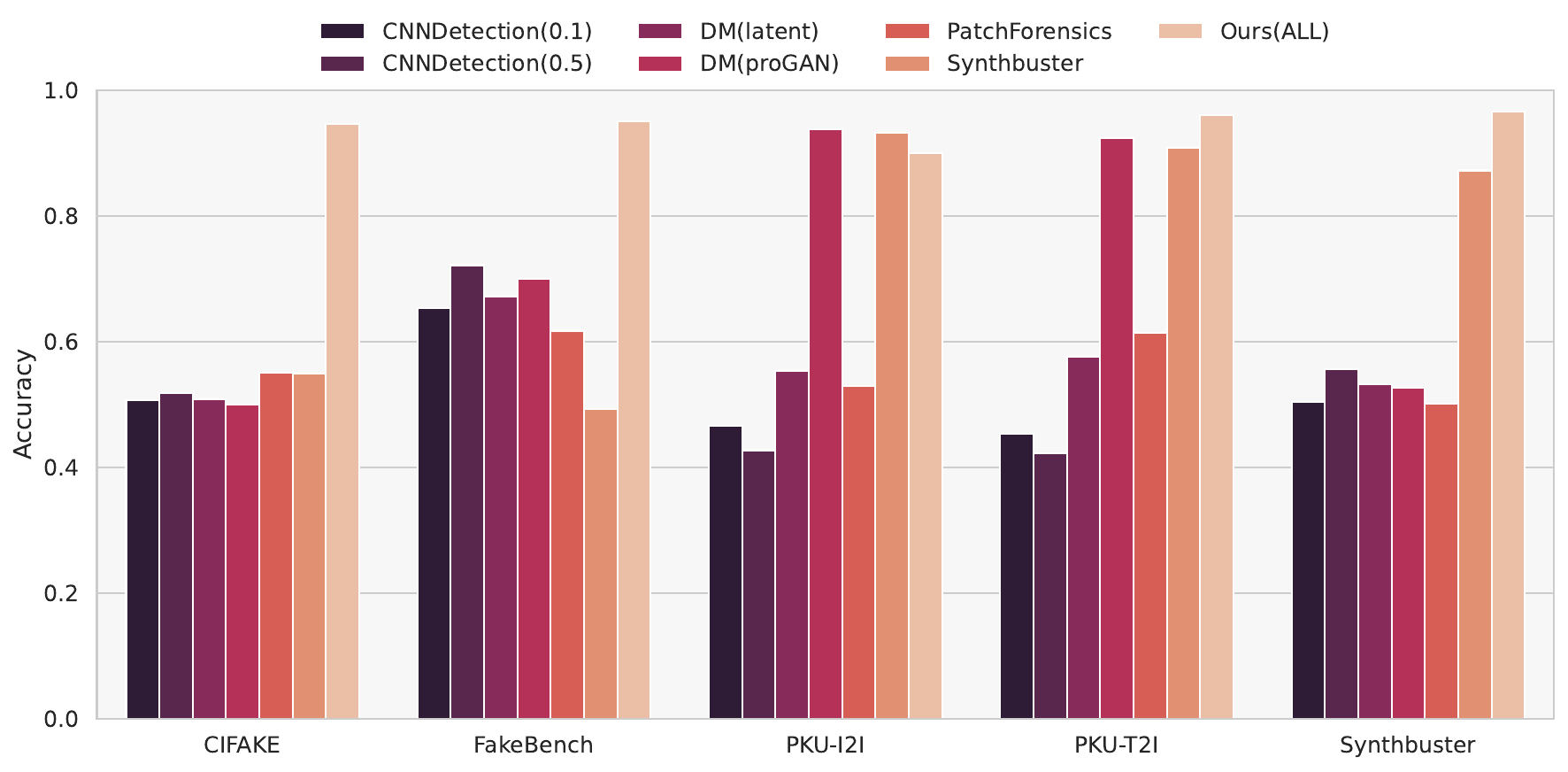}
    \caption{Performance comparison with state-of-the-art approaches.}
    \label{fig:sota_accuracy}
\end{figure}

% The quantitative comparison is summarized across the four datasets covering different generative scenarios.
Figure~\ref{fig:sota_accuracy} reports the Accuracy across datasets. The proposed framework consistently outperforms approaches, achieving an Accuracy of 0.946 on CIFAKE, 0.950 on FakeBench, 0.900 on PKU-I2I, 0.960 on PKU-T2I, and 0.966 on Synthbuster.

In contrast, existing approaches exhibit strong dataset dependency and limited cross-domain robustness. CNNDetection, trained primarily for GAN-based artifacts, performs near to chance level on several datasets, particularly CIFAKE and PKU-4k where the accuracy remains around $0.45$–$0.72$. Similarly, DMImageDetection shows inconsistent behavior, performing in specific configurations (ProGAN on PKU-4K, up to 0.938 Accuracy), but degrading significantly elsewhere. PatchForensics, which relies on local forensic inconsistencies, demonstrates slightly more stable, but modest performances, with accuracies between $0.50$ and $0.62$ across datasets.

Synthbuster performs well on its native benchmark (0.872 on Synthbuster) and reasonably on PKU-4K, but fails to generalize to more diverse settings such as FakeBench (0.492). These results highlight the difficulty of designing detectors that generalize across heterogeneous generative pipelines.

By contrast, the proposed approach maintains high performacnes across all datasets, demonstrating strong cross-domain robustness. This can be attributed to the fusion of complementary cues: semantic embeddings capture high-level inconsistencies, while statistical and texture features detect properties. Unlike existing methods that focus on specific artifact families, the proposed framework integrates multiple levels of representation, enabling superios generalization across generators and domains.

% representations and statistical texture cues in the fused feature space. While most existing detectors specialize in specific artifact families (GAN fingerprints, diffusion noise patterns, or patch-level inconsistencies), the proposed framework captures both high-level semantic inconsistencies and low-level statistical deviations from natural image distributions. As a result, it achieves substantially better generalization across generators, synthesis pipelines, and dataset domains.

\vspace{-0.1cm}
\section{Conclusion}
\label{sec:conclusion}

This article presented a systematic investigation of multi-feature fusion for GenAI image detection. By combining statistical (MSCN), semantic (CLIP), and textural (MLBP) representations, we demonstrate that fusion yields a more robust and generalizable detector than single-feature approaches. Experiments across four benchmarks and seven generative models revealed that fusion consistently improves both Accuracy and MCC, achieving up to 96.6\% accuracy and 0.933 MCC in mixed-model settings. Our manifold-based interpretation and statistical analyses confirmed that detection is fundamentally representation-driven and that fusion stabilizes the feature space geometry, reducing sensitivity to both generator type and classifier choice. Compared to state-of-the-art methods, the proposed framework achieves consistently superior performance across diverse scenarios

Despite these strengths, the proposed framework relies on pre-trained feature extractors, which may inherit biases from their training data and incur computational overhead. Performance also remains limited for certain generators (\textit{e.g.} VQDM), suggesting room for improvement. Future work will explore (i) adaptive fusion strategies that weight features based on input characteristics, (ii) end-to-end learning of complementary representations, and (iii) extensions to video-based GenAI detection. Overall, hybrid multi-level approaches appear essential to keep pace with the rapid evolution of generative models.

\footnotesize
{
\bibliographystyle{IEEEtran}
\bibliography{refs}

@article{8977347,
  author  = {T. Karras and S. Laine and T. Aila},
  journal = {IEEE TPAMI},
  title   = {A Style-Based Generator Architecture for Generative Adversarial Networks},
  year    = {2021},
  volume  = {43},
  number  = {12},
  pages   = {4217-4228}
}

@inproceedings{pmlr-v139-ramesh21a,
  title     = {Zero-Shot Text-to-Image Generation},
  author    = {A. Ramesh and M. Pavlov and et al.},
  booktitle = {ICML},
  pages     = {8821--8831},
  year      = {2021},
  volume    = {139},
  address   = {Virtual}
}

@inproceedings{Rombach_2022_CVPR,
  author    = {R. Rombach and A. Blattmann and D. Lorenz and P. Esser and B. Ommer},
  title     = {High-Resolution Image Synthesis With Latent Diffusion Models},
  booktitle = {IEEE/CVFCVPR},
  month     = {June},
  year      = {2022},
  pages     = {10684-10695},
  address   = {New Orleans, LA, USA}
}

@article{ruderman1994statistics,
  title   = {The statistics of natural images},
  author  = {Ruderman, Daniel L},
  journal = {Network: Computation in Neural Systems},
  year    = {1994}
}

@article{krizhevsky2009learning,
  title     = {Learning multiple layers of features from tiny images},
  author    = {Krizhevsky, Alex and Hinton, Geoffrey and others},
  year      = {2009},
  publisher = {Toronto, ON, Canada}
}

@article{mittal2012brisque,
  title     = {No-reference image quality assessment in the spatial domain},
  author    = {Mittal, Anish and Moorthy, Anush Krishna and Bovik, Alan Conrad},
  journal   = {IEEE TIP},
  volume    = {21},
  number    = {12},
  pages     = {4695--4708},
  year      = {2012},
  publisher = {IEEE}
}

@inproceedings{radford2021clip,
  title     = {Learning Transferable Visual Models From Natural Language Supervision},
  author    = {Radford, Alec and Kim, Jong Wook and Hallacy, Chris and et al.},
  booktitle = {ICML},
  pages     = {8748--8763},
  year      = {2021},
  volume    = {139},
  address   = {Virtual}
}

@inproceedings{karras2018progan,
  title     = {Progressive growing of GANs for improved quality},
  author    = {Tero Karras and Timo Aila and Samuli Laine and Jaakko Lehtinen},
  booktitle = {ICLR},
  year      = {2018},
  address   = {Vancouver, Canada}
}

@inproceedings{karras2019stylegan,
  author    = {Karras, Tero and Laine, Samuli and Aila, Timo},
  title     = {A Style-Based Generator Architecture for Generative Adversarial Networks},
  booktitle = {IEEE/CVF CVPR},
  month     = {June},
  year      = {2019},
  address   = {Long Beach, CA, USA}
}

@inproceedings{patchforensics,
  title        = {What makes fake images detectable? understanding properties that generalize},
  author       = {Chai, Lucy and Bau, David and Lim, Ser-Nam and Isola, Phillip},
  booktitle    = {ECCV},
  pages        = {103--120},
  year         = {2020},
  organization = {Springer},
  address      = {Glasgow, UK}
}

@inproceedings{ho2020ddpm,
  author    = {Ho, Jonathan and Jain, Ajay and Abbeel, Pieter},
  booktitle = {NeurIPS},
  pages     = {6840--6851},
  title     = {Denoising Diffusion Probabilistic Models},
  volume    = {33},
  year      = {2020},
  address   = {Virtual}
}

@inproceedings{rombach2022ldm,
  author    = {Rombach, Robin and Blattmann, Andreas and et al.},
  title     = {High-Resolution Image Synthesis With Latent Diffusion Models},
  booktitle = {IEEE/CVF},
  month     = {June},
  year      = {2022},
  pages     = {10684-10695},
  address   = {Virtual}
}

@inproceedings{wang2019cnngenerated,
  title     = {CNN-generated images are surprisingly easy to spot...for now},
  author    = {Wang, Sheng-Yu and Wang, Oliver and Zhang, Richard and Owens, Andrew and Efros, Alexei A},
  booktitle = {CVPR},
  year      = {2020},
  address   = {Virtual}
}

@article{11124461,
  author   = {Li, Yixuan and Liu, Xuelin and Wang, Xiaoyang and Lee, Bu Sung and Wang, Shiqi and Rocha, Anderson and Lin, Weisi},
  journal  = {IEEE TIFS},
  title    = {FakeBench: Probing Explainable Fake Image Detection via Large Multimodal Models},
  year     = {2025},
  volume   = {20},
  number   = {},
  pages    = {8730-8745},
  doi      = {10.1109/TIFS.2025.3597211}
}

@inproceedings{StealthDiffusion,
  author    = {Z. Zhou and K. Sun and Z. Chen and H. Kuang and X. Sun and R. Ji},
  title     = {StealthDiffusion: Towards Evading Diffusion Forensic Detection through Diffusion Model},
  year      = {2024},
  booktitle = {ACM ICM},
  pages     = {3627–3636},
  numpages  = {10},
  address   = {Melbourne VIC, Australia}
}

@article{10409290,
  author   = {Bird, Jordan J. and Lotfi, Ahmad},
  journal  = {IEEE Access},
  title    = {CIFAKE: Image Classification and Explainable Identification of AI-Generated Synthetic Images},
  year     = {2024},
  volume   = {12},
  number   = {},
  pages    = {15642-15650},
  doi      = {10.1109/ACCESS.2024.3356122}
}

@inproceedings{Cozzolino_2024_CVPR,
  author    = {D. Cozzolino and G. Poggi and R. Corvi and M. Nie{\ss}ner and L. Verdoliva},
  title     = {Raising the Bar of AI-generated Image Detection with {CLIP}},
  booktitle = {IEEE/CVF CVPR},
  month     = {June},
  year      = {2024},
  pages     = {4356-4366},
  address   = {Seattle, U.S.A}
}

@inproceedings{Durall_2020_CVPR,
  author    = {R. Durall and M. Keuper and J. Keuper},
  title     = {Watch Your Up-Convolution: {CNN Based} Generative Deep Neural Networks Are Failing to Reproduce Spectral Distributions},
  booktitle = {IEEE/CVF CVPR},
  month     = {June},
  year      = {2020},
  address   = {Virtual}
}

@article{10246417,
  author  = {Y. Ju and S. Jia and J. Cai and H. Guan and S. Lyu},
  journal = {IEEE TMM},
  title   = {{GLFF}: Global and Local Feature Fusion for AI-Synthesized Image Detection},
  year    = {2024},
  volume  = {26},
  number  = {},
  pages   = {4073-4085}
}

@inproceedings{Li_2024_CVPRW,
  author    = {Y. Li and Q. Bammey and M. Gardella and T. Nikoukhah and JM. Morel and M. Colom and RG. Von},
  title     = {MaskSim: Detection of Synthetic Images by Masked Spectrum Similarity Analysis},
  booktitle = {IEEE/CVF CVPR},
  month     = {June},
  year      = {2024},
  pages     = {3855-3865},
  address   = {Seattle, WA, USA}
}

@inproceedings{Yu_2024_CVPR,
  author    = {Z. Yu and J. Ni and Y. Lin and H. Deng and B. Li},
  title     = {DiffForensics: Leveraging Diffusion Prior to Image Forgery Detection and Localization},
  booktitle = {IEEE/CVF Conference on Computer Vision and Pattern Recognition},
  month     = {June},
  year      = {2024},
  pages     = {12765-12774},
  address   = {Seattle, WA, USA}
}

@inproceedings{Baraldi_2024_CoDE,
  title        = {Contrasting deepfakes diffusion via contrastive learning and global-local similarities},
  author       = {L. Baraldi and F. Cocchi and M. Cornia and L. Baraldi and A. Nicolosi and R. Cucchiara},
  booktitle    = {ECCV},
  pages        = {199--216},
  year         = {2024},
  organization = {Springer},
  address      = {Milan, Italy}
}

@inproceedings{Liu_2024_CVPR,
  author    = {H. Liu and Z. Tan and C. Tan and Y. Wei and J. Wang and Y. Zhao},
  title     = {Forgery-aware Adaptive Transformer for Generalizable Synthetic Image Detection},
  booktitle = {IEEE/CVF CVPR},
  month     = {June},
  year      = {2024},
  pages     = {10770-10780},
  address   = {Seattle, WA, USA}
}

@inproceedings{8638330,
  author    = {F. Matern and C. Riess and M. Stamminger},
  booktitle = {IEEE WACVw},
  title     = {Exploiting Visual Artifacts to Expose Deepfakes and Face Manipulations},
  year      = {2019},
  volume    = {},
  number    = {},
  pages     = {83-92},
  address   = {Waikoloa, HI, USA}
}

@article{6272356,
  author  = {A. Mittal and Ak. Moorthy and AC. Bovik},
  journal = {IEEE TIP},
  title   = {No-Reference Image Quality Assessment in the Spatial Domain},
  year    = {2012},
  volume  = {21},
  number  = {12},
  pages   = {4695-4708}
}

@article{6353522,
  author  = {A. Mittal and R. Soundararajan and AC. Bovik},
  journal = {IEEE SPL},
  title   = {Making a “Completely Blind” Image Quality Analyzer},
  year    = {2013},
  volume  = {20},
  number  = {3},
  pages   = {209-212}
}

@inproceedings{pmlr-v139-radford21a,
  title     = {Learning Transferable Visual Models From Natural Language Supervision},
  author    = {A. Radford and et al.},
  booktitle = {International Conference on Machine Learning},
  pages     = {8748--8763},
  year      = {2021},
  volume    = {139}
}

@article{10334046,
  author  = {Q. Bammey},
  journal = {IEEE Open Journal of Signal Processing},
  title   = {Synthbuster: Towards Detection of Diffusion Model Generated Images},
  year    = {2024},
  volume  = {5},
  number  = {},
  pages   = {1-9}
}

@inproceedings{dang2015raise,
  title     = {Raise: A raw images dataset for digital image forensics},
  author    = {D. Dang-Nguyen and C. Pasquini and V. Conotter and G. Boato},
  booktitle = {ACM International Workshop on Multimedia Security and Content Protection},
  pages     = {219--224},
  year      = {2015},
  address   = {Portland, U.S.A}
}

@article{baldi2000assessing,
  title     = {Assessing the accuracy of prediction algorithms for classification: an overview},
  author    = {Baldi, Pierre and Brunak, S{\o}ren and Chauvin, Yves and Andersen, Claus AF and Nielsen, Henrik},
  journal   = {Bioinformatics},
  volume    = {16},
  number    = {5},
  pages     = {412--424},
  year      = {2000},
  publisher = {Oxford University Press}
}

@article{Ruppert01062004,
  author  = {David Ruppert},
  title   = {The Elements of Statistical Learning: Data Mining, Inference, and Prediction},
  journal = {Journal of ASA},
  volume  = {99},
  number  = {466},
  pages   = {567--567},
  year    = {2004}
}

@article{benjamini1995controlling,
  title     = {Controlling the false discovery rate: a practical and powerful approach to multiple testing},
  author    = {Benjamini, Yoav and Hochberg, Yosef},
  journal   = {Journal of the Royal statistical society},
  volume    = {57},
  number    = {1},
  pages     = {289--300},
  year      = {1995},
  publisher = {Wiley Online Library}
}

@inproceedings{10095167,
  author    = {R. Corvi and D. Cozzolino and et al.},
  booktitle = {IEEE ICASSP},
  title     = {On The Detection of Synthetic Images Generated by Diffusion Models},
  year      = {2023},
  volume    = {},
  number    = {},
  pages     = {1-5},
  address   = {Rhodes Island, Greece}
}

@inproceedings{Yuan_2025_ICCV,
  author    = {Yuan, Jiquan and Li, Jihe and et al.},
  title     = {PKU-AIGIQA-4K: A Perceptual Quality Assessment Database for Both Text-to-Image and Image-to-Image AI-Generated Images},
  booktitle = {IEEE/CVF ICCVw},
  month     = {October},
  year      = {2025},
  pages     = {3331-3340},
  address   = {Honolulu, HI, USA}
}

@article{zhu2024genimage,
  title   = {Genimage: A million-scale benchmark for detecting ai-generated image},
  author  = {M. Zhu and et al.},
  journal = {ANIPS},
  volume  = {36},
  year    = {2024}
}

@article{lin2024detecting,
  title   = {Detecting multimedia generated by large ai models: A survey},
  author  = {Lin, Li and Gupta, Neeraj and Zhang, Yue and Ren, Hainan and Liu, Chun-Hao and Ding, Feng and Wang, Xin and Li, Xin and Verdoliva, Luisa and Hu, Shu},
  journal = {arXiv preprint arXiv:2402.00045
             
             
             
             
             
             
             
             
             
             
             
             },
  year    = {2024}
}

@article{pietikainen2010local,
  title   = {Local binary patterns},
  author  = {M. Pietik{\"a}inen},
  journal = {Scholarpedia},
  volume  = {5},
  number  = {3},
  pages   = {9775},
  year    = {2010}
}

@article{ricker2022towards,
  title   = {Towards the detection of diffusion model deepfakes},
  author  = {J. Ricker and S. Damm and T. Holz and A. Fischer},
  journal = {arXiv preprint arXiv:2210.14571
             
             
             
             },
  year    = {2022}
}

@inproceedings{schinas2024sidbench,
  title     = {SIDBench: A Python framework for reliably assessing synthetic image detection methods},
  author    = {M. Schinas and S. Papadopoulos},
  booktitle = {ACM International Workshop on Multimedia AI against Disinformation},
  pages     = {55--64},
  year      = {2024},
  address   = {Melbourne, Australia}
}
}

\vfill

\end{document}